%% file: main.tex
\documentclass[twoside]{article}

\usepackage[margin=1in]{geometry} 

\usepackage{authblk}

\usepackage{graphicx}
\usepackage{caption, subcaption}

\usepackage[utf8]{inputenc} 
\usepackage[T1]{fontenc}    
\usepackage[colorlinks=true,
            linkcolor=blue,
            urlcolor=blue,
            citecolor=blue]{hyperref}       
\usepackage{url}            
\usepackage{booktabs}       
\usepackage{amsfonts}       
\usepackage{nicefrac}       
\usepackage{microtype}      
\usepackage[table]{xcolor}         
\usepackage{enumitem}

\usepackage{mathtools}
\usepackage{amsmath}
\usepackage{amssymb}
\usepackage{enumitem}
\usepackage[mode=build]{standalone}
\usepackage{pdfpages}
\usepackage{float}
\usepackage{amsthm}
\usepackage{pgf}
\usepackage{placeins}
\usepackage{makecell}
\usepackage{natbib}

\usepackage{definitions}

\usepackage[capitalize,noabbrev]{cleveref}

\theoremstyle{plain}
\newtheorem{theorem}{Theorem}[section]
\newtheorem{proposition}[theorem]{Proposition}

\theoremstyle{definition}
\newtheorem{definition}[theorem]{Definition}
\newtheorem{assumption}[theorem]{Assumption}
\theoremstyle{remark}

\usepackage{svg}
\usepackage{pgf, tikz}
\usetikzlibrary{arrows.meta, automata, positioning}

\usepackage{array}
\usepackage{multicol}
\usepackage{multirow}

\usepackage{pifont}
\newcommand{\cmark}{\ding{51}}%
\newcommand{\xmark}{\ding{55}}%

\begin{document}
\title{Causally Inspired Regularization Enables Domain General Representations}
\date{} 					

\author[1]{Olawale Salaudeen\thanks{Contact: oes2@illinois.edu}}
\author[2]{Sanmi Koyejo}

\affil[1]{University of Illinois at Urbana Champaign}
\affil[2]{Stanford University}

\maketitle

\begin{abstract}
Given a causal graph representing the data-generating process shared across different domains/distributions, enforcing sufficient graph-implied conditional independencies can identify domain-general (non-spurious) feature representations. For the standard input-output predictive setting, we categorize the set of graphs considered in the literature into two distinct groups: (i) those in which the empirical risk minimizer across training domains gives domain-general representations and (ii) those where it does not. For the latter case (ii), we propose a novel framework with regularizations, which we demonstrate are sufficient for identifying domain-general feature representations without a priori knowledge (or proxies) of the spurious features. Empirically, our proposed method is effective for both (semi) synthetic and real-world data, outperforming other state-of-the-art methods in average and worst-domain transfer accuracy.\\
\end{abstract}

\input{00_paper/01_Introduction}
\input{00_paper/02_Related_Work}
\input{00_paper/03_Problem_Setup}

\input{00_paper/04_Objective}

\input{00_paper/06_Experiments}

\input{00_paper/07_Conclusion}
\bibliographystyle{plainnat}
\bibliography{main}

\newpage
\appendix
\input{00_paper/99_appendix_expr}
\FloatBarrier
\input{00_paper/99_appendix_theory}

\end{document}

%% file: 00_paper/01_Introduction.tex
\section{Introduction}\label{sec:intro}
A key feature of machine learning is its capacity to generalize across new domains. When these domains present different data distributions, the algorithm must leverage shared structural concepts to achieve out-of-distribution (OOD) or out-of-domain generalization. This capability is vital in numerous important real-world machine learning applications. For example, in safety-critical settings such as autonomous driving, a lack of resilience to unfamiliar distributions could lead to human casualties. Likewise, in the healthcare sector, where ethical considerations are critical, an inability to adjust to shifts in data distribution can result in unfair biases, manifesting as inconsistent performance across different demographic groups.

An influential approach to domain generalization is Invariant Causal Prediction (ICP; \citep{Peters2015CausalIU}). ICP posits that although some aspects of data distributions (like spurious or non-causal mechanisms \citep{Pearl2010CausalI}) may change across domains, certain causal mechanisms remain constant. ICP suggests focusing on these invariant mechanisms for prediction. However, the estimation method for these invariant mechanisms suggested by \citep{Peters2015CausalIU} struggles with scalability in high-dimensional feature spaces. To overcome this, \cite{Arjovsky2019InvariantRM} introduced Invariant Risk Minimization (IRM), designed to identify these invariant mechanisms by minimizing an objective. However, requires strong assumptions for identifying the desired domain-general solutions \citep{ahuja2021invariance, Rosenfeld2021AnOL}; for instance, observing a number of domains proportional to the spurious features' dimensions is necessary, posing a significant challenge in these high-dimensional settings.
 
Subsequent variants of IRM have been developed with improved capabilities for identifying domain-general solutions \citep{ahuja2020invariant, krueger2021out, robey2021model, wang2022provable, ahuja2021invariance}. Additionally, regularizers for Distributionally Robust Optimization with subgroup shift have been proposed (GroupDRO) \citep{sagawa2019distributionally}. However, despite their solid theoretical motivation, empirical evidence suggests that these methods may not consistently deliver domain-general solutions in practice \cite{Gul2020LostDG, kaur2022modeling, Rosenfeld2021AnOL}.

\cite{kaur2022modeling} demonstrated that regularizing directly for conditional independencies implied by the generative process can give domain-general solutions, including conditional independencies beyond those considered by IRM. However, their experimental approach involves regularization terms that require direct observation of spurious features, a condition not always feasible in real-world applications. Our proposed methodology also leverages regularizers inspired by the conditional independencies indicated by causal graphs but, crucially, it does so without necessitating prior knowledge (or proxies) of the spurious features.

\subsection{Contributions}
In this work,
\begin{itemize}
\item we outline sufficient properties to uniquely identify domain-general predictors for a general set of generative processes that include domain-correlated spurious features,
\item we propose regularizers to implement these constraints without independent observations of the spurious features, and
\item finally, we show that the proposed framework outperforms the state-of-the-art on semi-synthetic and real-world data.
\end{itemize}

\noindent The code for our proposed method is provided at \href{https://github.com/olawalesalaudeen/tcri}{https://github.com/olawalesalaudeen/tcri}.

\paragraph{Notation:} Capital letters denote bounded random variables, and corresponding lowercase letters denote their value. Unless otherwise stated, we represent latent domain-general features as $\Zc \in \Zcalc \equiv \RR^m$ and spurious latent features as $\Ze \in \Zcale \equiv \RR^o$. Let $X \in \mathcal{X} \equiv \RR^d$ be the observed feature space and the output space of an invertible function $\Gamma: \Zcalc \times \Zcale \mapsto \Xcal$ and $Y \in \mathcal{Y} \equiv \{0, 1, \ldots, K-1\}$ be the observed label space for a $K$-class classification task. We then define feature extractors aimed at identifying latent features $\Phic: \mathcal{X} \mapsto \mathbb{R}^{m}$, $\Phie: \mathcal{X} \mapsto \mathbb{R}^{o}$ so that $\Phi: \mathcal{X} \mapsto \mathbb{R}^{m+o} \,\big(\text{that is }\Phi(x) = [\Phic(x); \Phie(x)] \forall x \in \Xcal\big)$. We define $e$ as a discrete random variable denoting domains and $\Ecal = \{P^e(\Zc, \Ze, X, Y): e=1, 2, \ldots\}$ to be the set of possible domains. $\Etr \subset \Ecal$ is the set of observed domains available during training.

%% file: 00_paper/02_Related_Work.tex
\section{Related Work}
The source of distribution shift can be isolated to components of the joint distribution. One special case of distribution shift is {\em covariate shift} \citep{shimodaira2000improving, zadrozny2004learning, huang2006correcting, gretton2009covariate, sugiyama2007direct, bickel2009discriminative, chen2016robust, schneider2020improving}, where only the covariate distribution $P(X)$ changes across domains. \cite{BenDavid2009ATO} give upper-bounds on target error based on the $\mathcal{H}$-divergence between the source and target covariate distributions, which motivates domain alignment methods like the Domain Adversarial Neural Networks \citep{ganin2016domain} and others \citep{Long2015LearningTF, blanchard2017domain}. Others have followed up on this work with other notions of covariate distance for domain adaptation, such as mean maximum discrepancy (MMD) \citep{long2016unsupervised}, Wasserstein distance \citep{courty2017joint}, etc. However,  \cite{pmlr-v75-kpotufe18a} show that these divergence metrics fail to capture many important properties of transferability, such as asymmetry and non-overlapping support. Furthermore, \cite{Zhao2019OnLI} shows that even with the alignment of covariates, large distances between label distributions can inhibit transfer; they propose a label conditional importance weighting adjustment to address this limitation. Other works have also proposed conditional covariate alignment \citep{Combes2020DomainAW, Li2018DeepDG, li2018mmd}.

Another form of distribution shift is {\em label shift}, where only the label distribution changes across domains. \cite{Lipton2018DetectingAC} propose a method to address this scenario. \cite{schrouff2022maintaining} illustrate that many real-world problems exhibit more complex 'compound' shifts than just covariate or label shifts alone.

One can leverage {\em domain adaptation} to address distribution shifts; however, these methods are contingent on having access to unlabeled or partially labeled samples from the target domain during training. When such samples are available, more sophisticated domain adaptation strategies aim to leverage and adapt spurious feature information to enhance performance \citep{liu2021just, zhang2021adaptive, kirichenko2022last}. However, domain generalization, as a problem, does not assume access to such samples \citep{muandet2013domain}.

To address the domain generalization problem, Invariant Causal Predictors (ICP) leverage shared causal structure to learn domain-general predictors \citep{Peters2015CausalIU}.
Previous works, enumerated in the introduction (Section~\ref{sec:intro}), have proposed various algorithms to identify domain-general predictors. \cite{Arjovsky2019InvariantRM}'s proposed invariance risk minimization (IRM) and its variants motivated by domain invariance:
\begin{align*}\min_{w, \Phi} \frac{1}{|\Etr|}\sum_{e\in\Etr}R^e(w \circ \Phi) \text{ s.t. } w \in \argmin_{\tilde{w}} R^e(\tilde{w} \cdot \Phi), \forall e \in \Etr,\end{align*}
where $R^e(w \circ \Phi) = \EE\big[\ell(y, w\cdot\Phi(x))\big]$, with loss function $\ell$, feature extractor $\Phi$, and linear predictor $w$. This objective aims to learn a representation $\Phi$ such that predictor $w$ that minimizes empirical risks on average across all domains also minimizes within-domain empirical risk for all domains.
However, \cite{Rosenfeld2021TheRO, ahuja2020invariant} showed that this objective requires unreasonable constraints on the number of observed domains at train times, e.g., observing distinct domains on the order of the rank of spurious features. Follow-up works have attempted to improve these limitations with stronger constraints on the problem -- enumerated in the introduction section.

Our method falls under domain generalization; however, unlike the domain-general solutions previously discussed, our proposed solution leverages different conditions than domain invariance directly, which we show may be more suited to learning domain-general representations.

%% file: 00_paper/03_Problem_Setup.tex
\section{Causality and Domain Generalization} \label{sec:background}
We often represent causal relationships with a causal graph. A {\em causal graph} is a directed acyclic graph (DAG), $G = (V, E)$, with nodes $V$ representing random variables and directed edges $E$ representing causal relationships, i.e., parents are causes and children are effects. A {\em structural equation model} (SEM) provides a mathematical representation of the causal relationships in its corresponding DAG. Each variable $Y \in V$ is given by $Y = f_Y(X) + \varepsilon_Y$, where $X$ denotes the parents of $Y$ in $G$, $f_Y$ is a deterministic function, and $\varepsilon_Y$ is an error capturing exogenous influences on $Y$. The main property we need here is that $f_Y$ is invariant to {\em interventions} to $V\backslash\{Y\}$ and is consequently invariant to changes in $P(V)$ induced by these interventions. Interventions refer to changes to $f_Z$, $Z \in V\backslash\{Y\}$.

In this work, we focus on domain-general predictors $d_g$ that are linear functions of features with domain-general mechanisms, denoted as $\gc \coloneqq w \circ \Phic$, where $w$ is a linear predictor and $\Phic$ identifies features with domain-general mechanisms. We use domain-general rather than domain-invariant since domain-invariance is strongly tied to the property: ${Y \indep e \,|\, \Zc}$~\citep{Arjovsky2019InvariantRM}. As shown in the subsequent sections, this work leverages other properties of appropriate causal graphs to obtain domain-general features. This distinction is crucial given the challenges associated with learning domain-general features through domain-invariance methods~\citep{Rosenfeld2021TheRO}.

Given the presence of a distribution shift, it's essential to identify some common structure across domains that can be utilized for out-of-distribution (OOD) generalization. For example, \cite{shimodaira2000improving} assume $P(Y|X)$ is shared across all domains for the covariate shift problem. In this work, we consider a setting where each domain is composed of observed features and labels, $X\in\Xcal, Y\in\Ycal$, where $X$ is given by an invertible function $\Gamma$ of two latent random variables: domain-general $\Zc \in \Zcalc$ and spurious $\Ze \in \Zcale$. By construction, the conditional expectation of the label $Y$ given the domain-general features $\Zc$ is the same across domains, i.e.,
\begin{align}\label{eq:zc_inv}\EE_{e_i}\left[Y | \Zc=\zc\right] = \EE_{e_j}\left[Y | \Zc=\zc\right] \\ \forall \zc \in \Zcalc, \forall e_i \ne e_j \in \Ecal.\nonumber \end{align}
Conversely, this robustness to $e$ does not necessarily extend to spurious features $\Ze$; in other words, {\em$\Ze$ may assume values that could lead a predictor relying on it to experience arbitrarily high error rates.} Then, a sound strategy for learning a domain-general predictor -- one that is robust to distribution shifts -- is to identify the latent domain-general $\Zc$ from the observed features $X$.

The approach we take to do this is motivated by the {\em Reichenbach Common Cause Principle}, which claims that if two events are correlated, there is either a causal connection between the correlated events that is responsible for the correlation or there is a third event, a so-called (Reichenbachian) common cause, which brings about the correlation \citep{sep-physics-Rpcc, Redei2002}. This principle allows us to posit the class of generative processes or causal mechanisms that give rise to the correlated observed features and labels, where the observed features are a function of domain-general and spurious features. We represent these generative processes as causal graphs. Importantly, {\em the mapping from a node's causal parents to itself is preserved in all distributions generated by the causal graph} (Equation \ref{eq:zc_inv}), and distributions can vary arbitrarily so long as they preserve the conditional independencies implied by the DAG (Markov Property \citep{Pearl2010CausalI}).

We now enumerate DAGs that give observe features with spurious correlations with the label.

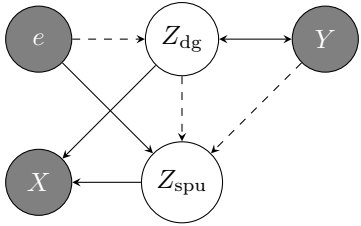
\begin{figure}[t]
    \centering
    \input{files/graph_no_ey_x}
    \caption{Partial Ancestral Graph representing all non-trivial and valid generative processes (DAGs); dashed edges indicate that an edge may or may not exist. \label{fig:pagb}}
    \label{fig:pag}
\end{figure}

\paragraph{Valid DAGs.} We consider generative processes, where both latent features, $\Ze, \Zc, \text{ and observed } X$ are correlated with $Y$, and the observed $X$ is a function of only $\Zc$ and $\Ze$ (Figure \ref{fig:pag}).

Given this setup, there is an enumerable set of valid generative processes. Such processes are (i) without cycles, (ii) are feature complete -- including edges from $\Zc$ and $\Ze$ to $X$, i.e., $\Zc \rightarrow X \leftarrow \Ze$, and (iii) where the observed features mediate domain influence, i.e., there is no direct domain influence on the label $e \not\rightarrow Y$. We discuss this enumeration in detail in Appendix \ref{sec:app_dags}. The result of our analysis is identifying a representative set of DAGs that describe valid generative processes -- these DAGs come from orienting the partial ancestral graph (PAG) in Figure \ref{fig:pagb}. We compare the conditional independencies implied by the DAGs defined by Figure \ref{fig:pagb} as illustrated in Figure \ref{fig:graph_other}, resulting in three canonical DAGs in the literature (see Appendix \ref{sec:app_dags} for further discussion). Other DAGs that induce spurious correlations are outside the scope of this work.

\begin{figure}[h]
    \centering
    \begin{subfigure}[t]{0.31\textwidth}
        \resizebox{\textwidth}{!}{
            \input{files/graph_1.tex}
        }
    \caption{Causal \citep{Arjovsky2019InvariantRM}.} \label{fig:first}
    \end{subfigure}
    \hfill
    \begin{subfigure}[t]{0.31\textwidth}
        \resizebox{\textwidth}{!}{
            \input{files/graph_2.tex}
        }
    \caption{Anticausal \citep{Rosenfeld2021TheRO}.} \label{fig:second}
    \end{subfigure}
    \hfill
    \begin{subfigure}[t]{0.31\textwidth}
        \resizebox{\textwidth}{!}{
            \input{files/graph_3.tex}
        }
    \caption{Fully Informative Causal \citep{ahuja2021invariance}.} \label{fig:third}
    \end{subfigure}
    \caption{{\bf Generative Processes.} Graphical models depicting the structure of possible data-generating processes -- shaded nodes indicate observed variables. $X$ represents the observed features, $Y$ represents observed targets, and $e$ represents domain influences (domain indexes in practice). There is an explicit separation of domain-general $\Zc$ and domain-specific $\Ze$ features; they are combined to generate observed $X$. Dashed edges indicate the possibility of an edge.}
    \label{fig:graph_other}
\end{figure}
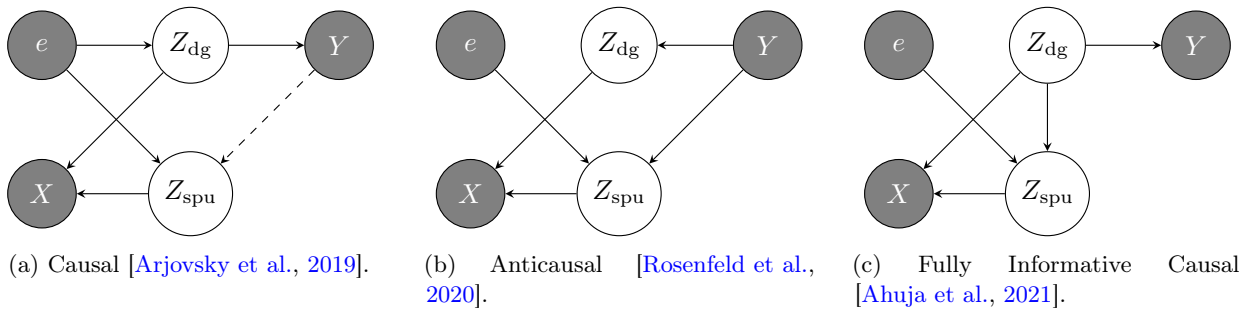

\paragraph{Conditional independencies implied by identified DAGs (Figure~\ref{fig:graph_other}).}
\begin{enumerate}[label=(\alph*), align=left]
    \item [{\bf Fig.~\ref{fig:first}:}] $\bf{\Zc \indep \Ze \,|\, \{Y, e\}}$; $Y \indep e \,|\, \Zc$. \\
 
    This causal graphical model implies that the mapping from $\Zc$ to its causal child $Y$ is preserved and consequently, Equation \ref{eq:zc_inv} holds \citep{Pearl2010CausalI, Peters2015CausalIU}. As an example, consider the task of predicting the spread of a disease. Features may include causes (vaccination rate and public health policies) and effects (coughing). $e$ is the time of month; the distribution of coughing changes depending on the season.

    \item [{\bf Fig.~\ref{fig:second}:}] ${\bf\Zc \indep \Ze \,|\, \{Y, e\}}$; $\Zc \indep \Ze \,|\, Y$; $Y \indep e \,|\, \Zc$, $\Zc \indep e$.\\
 
    The causal graphical model does not directly imply that $\Zc \rightarrow Y$ is preserved across domains. However, in this work, it represents the setting where the inverse of the causal direction is preserved (inverse: $\Zc \rightarrow Y$), and thus Equation \ref{eq:zc_inv} holds. A context where this setting is relevant is in healthcare where medical conditions ($Y$) cause symptoms ($\Zc$), but the prediction task is often predicting conditions from symptoms, and this mapping $\Zc \rightarrow Y$, opposite of the causal direction, is preserved across distributions. Again, we may consider $e$ as the time of month; the distribution of coughing changes depending on the season.
    \item [{\bf Fig.~\ref{fig:third}:}]$Y \indep e \,|\, \Zc$; $\Zc \indep e$. \\
     
    Similar to Figure~\ref{fig:first}, this causal graphical model implies that the mapping from $\Zc$ to its causal child $Y$ is preserved, so Equation \ref{eq:zc_inv} holds \citep{Pearl2010CausalI, Peters2015CausalIU}. This setting is especially interesting because it represents a \textbf{Fully Informative Invariant Features setting}, that is $\Ze \indep Y \,|\, \Zc$ \citep{ahuja2021invariance}. Said differently, $\Ze$ does not induce a backdoor path from $e$ to $Y$ that $\Zc$ does not block. As an example of this, we can consider the task of predicting hospital readmission rates. Features may include the severity of illness, which is a direct cause of readmission rates, and also include the length of stay, which is also caused by the severity of illness. However, length of stay may not be a cause of readmission; the correlation between the two would be a result of the confounding effect of a common cause, illness severity. $e$ is an indicator for distinct hospitals. 
\end{enumerate}

\begin{table}[t]
    \centering
    \caption{Generative Processes and Sufficient Conditions for Domain-Generality}
    \renewcommand{\arraystretch}{1.1}
    \label{tab:easy_hard_DAGS}
    \resizebox{0.5\textwidth}{!}{
    \begin{tabular}{c|c|c|c}
        & \multicolumn{3}{c}{Graphs in Figure \ref{fig:graph_other}} \\
        \hline
        & (a) & (b) & (c) \\
        \hline
         $\Zc \indep \Ze \,|\, \{Y, e\}$ & \cmark & \cmark & \xmark \\
         Identifying $\Zc$ is necessary & \cmark & \cmark & \xmark \\
    \end{tabular}
}
\end{table}

We call the condition $\mathbf{Y \indep e \,|\, \Zc}$ the {\em domain invariance property}. This condition is common to all the DAGs in Figure~\ref{fig:graph_other}. We call the condition ${\Zc \indep \Ze \,|\, \{Y, e\}}$ the {\em target conditioned representation independence (TCRI) property.} This condition is  common to the DAGs in Figure~\ref{fig:first}, \ref{fig:second}. In the settings considered in this work, the TCRI property is equivalently $\mathbf{\Zc \indep \Ze \,|\, Y \forall e \in \Ecal}$ since $e$ will simply index the set of empirical distributions available at training.

\paragraph{Domain generalization with conditional independencies.} \cite{kaur2022modeling} showed that sufficiently regularizing for the correct conditional independencies described by the appropriate DAGs can give domain-general solutions, i.e., identifies $\Zc$. However, in practice, one does not (partially) observe the latent features independently to regularize directly. Other works have also highlighted the need to consider generative processes when designing robust algorithms to distribute shifts \citep{veitch2021counterfactual, pmlr-v151-makar22a}. However, previous work has largely focused on regularizing for the domain invariance property, ignoring the conditional independence property $\Zc \indep \Ze \,|\, Y, e$.
 
\paragraph{Sufficiency of ERM under Fully Informative Invariant Features.} Despite the known challenges of learning domain-general features from the domain-invariance properties in practice, this approach persists, likely due to it being the only property shared across all DAGs. We alleviate this constraint by observing that Graph (Fig.~\ref{fig:third}) falls under what \cite{ahuja2021invariance} refer to as the fully informative invariant features settings, meaning that $\Ze$ is redundant, having only information about $Y$ that is already in $\Zc$. \cite{ahuja2021invariance} show that the empirical risk minimizer is domain-general for bounded features.

\paragraph{Easy vs. hard DAGs imply the generality of TCRI.} Consequently, we categorize the generative processes into {\em easy} and {\em hard} cases Table \ref{tab:easy_hard_DAGS}: {\em (i) easy} meaning that minimizing average risk gives domain-general solutions, i.e., ERM is sufficient (Fig. \ref{fig:third}), and {\em (ii) hard} meaning that one needs to identify $\Zc$ to obtain domain-general solutions (Figs. \ref{fig:first}-\ref{fig:second}). We show empirically that regularizing for $\Zc \indep \Ze \,|\, Y \forall e \in \Ecal$ also gives a domain-general solution in the easy case.
{\bf The generality of TCRI} follows from its sufficiency for identifying domain-general $\Zc$ in the hard cases while still giving domain-general solutions empirically in the easy case.

\section{Proposed Learning Framework}
We have now clarified that hard DAGs (i.e., those not solved by ERM) share the TCRI property. The challenge is that $\Zc$ and $\Ze$ are not independently observed; otherwise, one could directly regularize. Existing work such as \cite{kaur2022modeling} empirically study semi-synthetic datasets where $\Ze$ is (partially) observed and directly learn $\Zc$ by regularizing that $\Phi(X) \indep \Ze \,|\, Y, e$ for feature extractor $\Phi$. To our knowledge, we are the first to leverage the TCRI property without requiring observation of $\Ze$. Next, we set up our approach with some key assumptions. The first is that the observed distributions are Markov to an appropriate DAG.

\begin{assumption} \label{assum:generative}
All distributions, sources and targets, are generated by one of the structural causal models $\mathcal{SCM}$ that follow:\\
%
\begin{minipage}{0.48\textwidth}
\begin{equation} \label{eq:sem}
    \overbrace{\mathcal{SCM}(e)}^{causal} \coloneqq
    \begin{cases}
      \Zc^{(e)} \sim P_{\Zc}^{(e)}, \\
      Y^{(e)} \leftarrow \langle \wc^*, \Zc^{(e)}\rangle + \eta_Y, \\
      \Ze^{(e)} \leftarrow \langle \we^*, Y\rangle + \eta_{\Ze}^{(e)}, \\
      X \leftarrow \Gamma(\Zc, \Ze),
    \end{cases}
\end{equation}
\end{minipage}
\hfill
\begin{minipage}{0.48\textwidth}
\begin{equation} \label{eq:sem2}
    \overbrace{\mathcal{SCM}(e)}^{anticausal} \coloneqq
    \begin{cases}
      Y^{(e)} \sim P_{Y}, \\
      \Zc^{(e)} \leftarrow \langle\wtc, Y\rangle + \eta_{\Zc}^{(e)}, \\
      \Ze^{(e)} \leftarrow \langle\we^*, Y\rangle + \eta_{\Ze}^{(e)}, \\
      X \leftarrow \Gamma(\Zc, \Ze),
    \end{cases}
\end{equation}
\end{minipage}
\begin{equation} \label{eq:sem3}
    \overbrace{\mathcal{SCM}(e)}^{FIIF} \coloneqq
    \begin{cases}
      \Zc^{(e)} \sim P_{\Zc}^{(e)}, \\
      Y^{(e)} \leftarrow \langle \wc^*, \Zc^{(e)}\rangle + \eta_Y, \\
      \Ze^{(e)} \leftarrow \langle \we^*, \Zc\rangle + \eta_{\Ze}^{(e)}, \\
      X \leftarrow \Gamma(\Zc, \Ze),
    \end{cases}
\end{equation}
~\\ where $P_{\Zc}$ is the causal covariate distribution, $w$'s are linear generative mechanisms, $\eta$'s are exogenous independent noise variables, and $\Gamma: \Zcalc \times \Zcale \rightarrow \Xcal$ is an invertible function. It follows from having causal mechanisms that we can learn a predictor $\wc^*$ for $\Zc$ that is domain-general (Equation \ref{eq:sem}-\ref{eq:sem3}) -- $\wc^*$ inverts the mapping $\wtc$ in the anticausal case.
\end{assumption}

These structural causal models (Equation \ref{eq:sem}-\ref{eq:sem3}) correspond to causal graphs Figures \ref{fig:first}-\ref{fig:third}, respectively.

\begin{assumption}[Structural] \label{assum:corr}
    Causal Graphs and their distributions are Markov and Faithful \citep{Pearl2010CausalI}.
\end{assumption}

Given Assumption \ref{assum:corr}, we aim to leverage TCRI property ($\Zc \indep \Ze \,|\, Y \forall e \in \Etr$) to learn the latent $\Zc$ without observing $\Ze$ directly. 
We do this by learning two feature extractors that, together, recover $\Zc$ and $\Ze$ and satisfy TCRI (Figure \ref{fig:architecture}). We formally define these properties as follows.

%% file: files/graph_no_ey_x.tex
\begin{tikzpicture}[node distance = 0.75in]
    \tikzstyle{every state}=[
        fill = white,
    ]

    \node[state, fill=gray, text=white] (E) {\normalsize $e$};
    \node[state, right of = E] (Zc) {\normalsize $\Zc$};
    \node[state, below of = Zc] (Ze) {\normalsize $\Ze$};
    \node[state, fill=gray, text=white, right of = Zc] (Y) {\normalsize $Y$};
    \node[state, fill=gray, text=white, below of = E] (X) {\normalsize $X$};

    \path[->, dashed, draw, >=stealth] (E) edge node {} (Zc);
    \path[->, draw, >=stealth] (E) edge node {} (Ze);
    \path[->, draw, >=stealth] (Zc) edge node {} (X);
    \path[->, dashed, draw, >=stealth] (Y) edge node {} (Ze);
    \path[<->, draw, >=stealth] (Zc) edge node[above] {} (Y);
    \path[->, dashed, draw, >=stealth] (Zc) edge node {} (Ze);
    \path[->, draw, >=stealth] (Ze) edge node {} (X);
\end{tikzpicture}

%% file: files/graph_1.tex
\begin{tikzpicture}[node distance = 0.75in]
    \tikzstyle{every state}=[
        fill = white,
    ]

    \node[state, fill=gray, text=white] (E) {\normalsize $e$};
    \node[state, right of = E] (Zc) {\normalsize $\Zc$};
    \node[state, below of = Zc] (Ze) {\normalsize $\Ze$};
    \node[state, fill=gray, text=white, right of = Zc] (Y) {\normalsize $Y$};
    \node[state, fill=gray, text=white, below of = E] (X) {\normalsize $X$};

    \path[->, draw, >=stealth] (E) edge node {} (Zc);
    \path[->, draw, >=stealth] (E) edge node {} (Ze);
    \path[->, draw, >=stealth] (Zc) edge node {} (X);
    \path[->, dashed, draw, >=stealth] (Y) edge node {} (Ze);
    \path[->, draw, >=stealth] (Zc) edge node {} (Y);
    \path[->, draw, >=stealth] (Ze) edge node {} (X);
\end{tikzpicture}

%% file: files/graph_2.tex
\begin{tikzpicture}[node distance = 0.75in]
    \tikzstyle{every state}=[
        fill = white,
    ]

    \node[state, fill=gray, text=white] (E) {\normalsize $e$};
    \node[state, right of = E] (Zc) {\normalsize $\Zc$};
    \node[state, below of = Zc] (Ze) {\normalsize $\Ze$};
    \node[state, fill=gray, text=white, right of = Zc] (Y) {\normalsize $Y$};
    \node[state, fill=gray, text=white, below of = E] (X) {\normalsize $X$};

    \path[->, draw, >=stealth] (E) edge node {} (Ze);
    \path[->, draw, >=stealth] (Zc) edge node {} (X);
    \path[->, draw, >=stealth] (Y) edge node {} (Ze);
    \path[->, draw, >=stealth] (Y) edge node {} (Zc);
    \path[->, draw, >=stealth] (Ze) edge node {} (X);
\end{tikzpicture}

%% file: files/graph_3.tex
\begin{tikzpicture}[node distance = 0.75in]
    \tikzstyle{every state}=[
        fill = white,
    ]

    \node[state, fill=gray, text=white] (E) {\normalsize $e$};
    \node[state, right of = E] (Zc) {\normalsize $\Zc$};
    \node[state, below of = Zc] (Ze) {\normalsize $\Ze$};
    \node[state, fill=gray, text=white, right of = Zc] (Y) {\normalsize $Y$};
    \node[state, fill=gray, text=white, below of = E] (X) {\normalsize $X$};

    \path[->, draw, >=stealth] (E) edge node {} (Ze);
    \path[->, draw, >=stealth] (Zc) edge node {} (X);
    \path[->, draw, >=stealth] (Zc) edge node {} (Ze);
    \path[->, draw, >=stealth] (Zc) edge node {} (Y);
    \path[->, draw, >=stealth] (Ze) edge node {} (X);
\end{tikzpicture}

%% file: 00_paper/04_Objective.tex
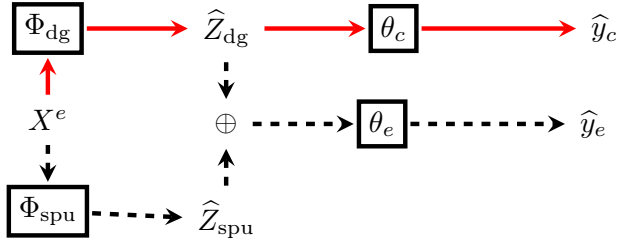
\begin{figure}[!t] 
    \centering
    \resizebox{0.5\textwidth}{!}{\input{files/architecture.tex}}
    \caption{{\bf Modeling approach.} During training, both representations, $\Phic$, and $\Phie$, generate domain-general and domain-specific predictions, respectively. However, only the domain-invariant representations/predictions are used during testing -- indicated by the solid red arrows.}
    \label{fig:architecture}
\end{figure}

\begin{definition}[Total Information Criterion (TIC)] \label{def:tcic}
$\Phi = \Phic \oplus \Phie$ satisfies TIC with respect to random variables $X,\, Y,\, e$ if for $\Phi(X^e) = [\Phic(X^e); \Phie(X^e)]$, there exists a linear operator $\Tcal$ s.t., $\Tcal(\Phi(X^e)) = [\Zc^e; \Ze^e] \forall e \in \Etr$.
\end{definition}
In other words, a feature extractor that satisfies the total information criterion recovers the complete latent feature sets $\Zc,\, \Ze$. This allows us to define the proposed implementation of the TCRI property non-trivially -- the conditional independence of subsets of the latents may not have the same implications on domain generalization. We note that $X \indep Y | \Zc, \Ze$, so $X$ has no information about $Y$ that is not in $\Zc, \Ze$.
\begin{definition}[Target Conditioned Representation Independence] \label{def:tcri}
$\Phi = \Phic \oplus \Phie$ satisfies TCRI with respect to random variables $X,\, Y,\, e$ if $\Phic(X) \indep \Phie(X) \,|\, Y \forall e \in \Ecal$.
\end{definition}
\begin{proposition} \label{prop:main} Assume that $\Phic(X)$ and $\Phie(X)$ are correlated with $Y$. Given Assumptions \ref{assum:generative}-\ref{assum:corr} and a representation $\Phi = \Phic \oplus \Phie$ that satisfies TIC, $\Phic(X) = \Zc \iff$ $\Phi$ satisfies TCRI. (see Appendix \ref{sec:app_theory} for proof).
\end{proposition}
Proposition \ref{prop:main} shows that TCRI is necessary and sufficient to identify $\Zc$ from a set of training domains. We note that we can verify if $\Phic(X)$ and $\Phie(X)$ are correlated with $Y$ by checking if the learned predictors are equivalent to chance.  Next, we describe our proposed algorithm to implement the conditions to learn such a feature map. Figure \ref{fig:architecture} illustrates the learning framework.

\paragraph{Learning Objective:} The first term in our proposed objective is $$\Lcal_{\Phic} = \mathcal{R}^e(\theta_c \circ \Phic),$$
where $\Phic: \mathcal{X} \mapsto \RR^m$ is a feature extractor, $\theta_c: \RR^{m} \mapsto \mathcal{Y}$ is a linear predictor, and $\Rcal^e(\theta_c\circ \Phic) = \EE\big[\ell(y, \theta_c\cdot\Phi(x))\big]$ is the empirical risk achieved by the feature extractor and predictor pair on samples from domain $e$. $\Phic$ and $\theta_c$ are designed to capture the domain-general portion of the framework.

Next, to implement the total information criterion, we use another feature extractor $\Phie: \mathcal{X} \mapsto \RR^o$, designed to capture the domain-specific information in $X$ that is not captured by $\Phic$. Together, we have $\Phi = \Phic \oplus \Phie$ where $\Phi$ has domain-specific predictors $\theta_e: \RR^{m+o} \mapsto \mathcal{Y}$ for each training domain, allowing the feature extractor to utilize domain-specific information to learn distinct optimal domain-specific (non-general) predictors:
\begin{equation*}
    \mathcal{L}_{\Phi} = \mathcal{R}^{e}\big( \theta_{e} \circ \Phi \big).
\end{equation*}
$\Lcal_{\Phi}$ aims to ensure that $\Phic \text{ and } \Phie$ capture all of the information about $Y$ in $X$ -- total information criterion. Since we do not know $o, m$, we select them to be the same size on our experiments; $o, m$ could be treated as hyperparameters though we do not treat them as such.

Finally, we implement the TCRI property (Definition \ref{def:tcri}). We denote $\mathcal{L}_{TCRI}$ to be a conditional independence penalty for $\Phic \text{ and } \Phie$. We utilize the Hilbert Schmidt independence Criterion (HSIC) \citep{Gretton2007AKS} as $\Lcal_{TCRI}$. However, in principle, any conditional independence penalty can be used in its place. \textbf{HSIC:}
\begin{align*}
    & \mathcal{L}_{TCRI}(\Phic, \Phie) = \frac{1}{2}\sum_{k\in\{0,1\}} \widehat{HSIC}\Big(\Phic(X), \Phie(X)\Big)^{y=k}
    = \frac{1}{2}\sum_{k\in\{0,1\}} \frac{1}{n_k^2}\text{tr}\Big(\mathbf{K}_{\Phic}\mathbf{H}_{n_k}\mathbf{K}_{\Phie}\mathbf{H}_{n_k}\Big)^{y=k},
\end{align*}
\noindent where $k$, indicates which class the examples in the estimate correspond to, $C$ is the number of classes, $\mathbf{K}_{\Phic}\in \mathbb{R}^{n_k\times n_k},\, \mathbf{K}_{\Phie}\in \mathbb{R}^{n_k\times n_k}$ are Gram matrices, $\mathbf{K}_{\Phi}^{i,j} = \kappa(\Phic(X)_i,\Phic(X)_j)$, $\mathbf{K}_{\Phie}^{i,j} = \omega(\Phie(X)_i, \Phie(X)_j)$ with kernels $\kappa, \omega$ are radial basis functions, $\mathbf{H}_{n_k} = \mathbf{I}_{n_k} - \frac{1}{n_k^2}\mathbf{1}\mathbf{1}^\top$ is a centering matrix, $\mathbf{I}_{n_k}$ is the ${n_k} \times {n_k}$ dimensional identity matrix, $\mathbf{1}_{n_k}$ is the ${n_k}$-dimensional vector whose elements are all 1, and $^\top$ denotes the transpose. We condition on the label by taking only examples of each label and computing the empirical HSIC; then, we take the average.

Taken together, the full objective to be minimized is as follows:
\begin{align}
    \begin{split}
    \mathcal{L} = \frac{1}{E_{tr}}\sum_{e\in\mathcal{E}_{tr}}\Bigg[ \mathcal{R}^e(\theta_c \circ \Phic) + 
    \mathcal{R}^e(\theta_{e} \circ \Phi)  + \beta\Lcal_{TCRI}(\Phic, \Phie) 
    \Bigg], \nonumber
    \end{split}
\end{align}
where $\beta > 0$ is a hyperparameter and $E_{tr}$ is the number of training domains. Figure \ref{fig:architecture} shows the full framework. We note that when $\beta=0$, this loss reduces to ERM.

Note that while we minimize this objective with respect to $\Phi, \theta_c, \theta_1, \ldots, \theta_{E_{tr}}$, only the domain-general representation and its predictor, $\theta_c \cdot \Phic$ are used for inference.

%% file: files/architecture.tex
\begin{tikzpicture}[node distance = 0.7in, >=stealth, line width=1.5pt]
    \tikzstyle{every state}=[anchor=west]

    \node[text=black] (x) {$X^{e}$};
    \node[rectangle, draw, text=black, right=0.5in of x, above=0.2in of x] (phi) {$\Phi_{\text{dg}}$};
    \node[rectangle, draw, text=black, right=0.5in of x, below=0.2in of x] (psi) {$\Phi_{\text{spu}}$};
    \node[text=black, right=0.5in of phi] (zc) {$\widehat{Z}_{\text{dg}}$};
    \node[rectangle, draw, text=black, right=0.5in of zc] (fc) {$\theta_c$};
    \node[text=black, below=0.2in of zc] (concat) {$\oplus$};
    \node[text=black, below=0.2in of concat] (ze) {$\widehat{Z}_{\text{spu}}$};
    \node[rectangle, draw, text=black, right=0.5in of concat] (fe) {$\theta_{e}$};
    \node[text=black, right=0.5in of fc, xshift=0.25in] (yc) {$\widehat{y}_{c}$};
    \node[text=black, right=0.5in of fe, xshift=0.25in] (ye) {$\widehat{y}_{e}$};
    
    \path[->, draw, color=red, shorten >=1pt, shorten <=1pt] (x) edge (phi);
    \path[->, dashed, draw, shorten >=1pt, shorten <=1pt] (x) edge (psi);
    \path[->, draw, color=red, shorten >=1pt, shorten <=1pt] (phi) edge (zc);
    \path[->, dashed, draw, shorten >=1pt, shorten <=1pt] (psi) edge (ze);
    \path[->, dashed, draw, shorten >=1pt, shorten <=1pt] (zc) edge (concat);
    \path[->, dashed, draw, shorten >=1pt, shorten <=1pt] (ze) edge (concat);
    \path[->, draw, color=red, shorten >=1pt, shorten <=1pt] (zc) edge (fc);
    \path[->, dashed, draw, shorten >=1pt, shorten <=1pt] (concat) edge (fe);
    \path[->, draw, color=red, shorten >=1pt, shorten <=1pt] (fc) edge (yc);
    \path[->, dashed, draw, shorten >=1pt, shorten <=1pt] (fe) edge (ye);
\end{tikzpicture}

%% file: 00_paper/06_Experiments.tex
\newcolumntype{?}{!{\vrule \vrule}}
\section{Experiments} \label{sec:tcri_expr}
We begin by evaluating with simulated data, i.e., with known ground truth mechanisms; we use Equation \ref{eq:simsem} to generate our simulated data, with domain parameter $\sigma_{e_i}$; code is provided in the supplemental materials.

\begin{table}[h]
\begin{minipage}{0.40\textwidth}
\begin{align} \label{eq:simsem}
    \mathcal{SCM}(e_i) \coloneqq
    \begin{cases}
      \Zc^{(e_i)} \sim \Ncal\left(0, \sigma_{e_i}^2\right)& \\
      y^{(e_i)} = \Zc^{(e_i)} + \Ncal\left(0, \sigma_y^2\right),\\
      \Ze^{(e_i)} = Y^{(e_i)} + \Ncal\left(0, \sigma_{e_i}^2\right).
    \end{cases} 
\end{align}
\end{minipage}
\hfill
\centering
\begin{minipage}{0.58\textwidth}
\centering
\caption{Continuous Simulated Results -- Feature Extractor with a dummy predictor $\theta_c = 1.$, i.e., $\hat{y} = x \cdot \Phic \cdot w$, where $x\in \mathbb{R}^{N \times 2},\, \Phic, \Phie \in \mathbb{R}^{2 \times 1},\, w \in \mathbb{R}$. Oracle indicates the coefficients achieved by regressing $y$ on $z_c$ directly.}
\label{tab:sim}
\input{results/simulated}
\end{minipage}
\end{table}

We observe 2 domains with parameters $\sigma_{e=0} = 0.1$, $\sigma^{e=1} = 0.2$ with $\sigma_y = 0.25$, 5000 samples, and linear feature extractors and predictors. We use partial covariance as our conditional independence penalty $\Lcal_{TCRI}$. Table ~\ref{tab:sim} shows the learned value of $\Phic$, where `Oracle' indicates the true coefficients obtained by regressing $Y$ on domain-general $\Zc$ directly. The ideal $\Phic$ recovers $\Zc$ and puts zero weight on $\Ze$. 

Now, we evaluate the efficacy of our proposed objective on non-simulated datasets.
 
\subsection{Semisynthetic and Real-World Datasets}

\paragraph{Algorithms:} We compare our method to baselines corresponding to DAG properties: Empirical Risk Minimization (\textbf{ERM}, \citep{NIPS1991_ff4d5fbb}), Invariant Risk Minimization (\textbf{IRM} \citep{Arjovsky2019InvariantRM}), Variance Risk Extrapolation (\textbf{V-REx}, \citep{krueger2021out}), \citep{li2018learning}), Group Distributionally Robust Optimization (\textbf{GroupDRO}), \citep{sagawa2019distributionally}), and Information Bottleneck methods (\textbf{IB\_ERM/IB\_IRM},  \citep{ahuja2021invariance}). Additional baseline methods are provided in the Appendix \ref{app:add_res}.

We evaluate our proposed method on the semisynthetic ColoredMNIST \citep{Arjovsky2019InvariantRM} and real-world Terra Incognita dataset \citep{beery2018recognition}. Given observed domains $\Etr = \{e: 1,2,\ldots,E_{tr}\}$, we train on $\mathcal{E}_{tr} \, \backslash \, e_i$ and evaluate the model on the unseen domain $e_i$, for each $e \in \Etr$.

\paragraph{\em ColoredMNIST:} The ColoredMNIST dataset \citep{Arjovsky2019InvariantRM} is composed of $7000$ ($2\times 28 \times 28$, $1$) images of a hand-written digit and binary-label pairs. There are three domains with different correlations between image color and label, i.e., the image color is spuriously related to the label by assigning a color to each of the two classes (0: digits 0-4, 1: digits 5-9). The color is then flipped with probabilities $\{0.1, 0.2, 0.9\}$ to create three domains, making the color-label relationship domain-specific because it changes across domains. There is also label flip noise of $0.25$, so we expect that the best accuracy a domain-general model can achieve is 75\%, while a non-domain general model can achieve higher. In this dataset, $\Zc$ corresponds to the original image, $\Ze$ the color, $e$ the label-color correlation, $Y$ the image label, and $X$ the observed colored image. This DAG follows the generative process of Figure \ref{fig:first} \citep{Arjovsky2019InvariantRM}.

\paragraph{\em Spurrious PACS:} \emph{Variables.} $X$: images, $Y$: non-urban (elephant, giraffe, horse) vs. urban (dog, guitar, house, person). \emph{Domains.} \{\{cartoon,  art painting\}, \{art painting,  cartoon\}, \{photo\}\} \citep{li2017deeper}. The photo domain is the same as in the original dataset. In the \{cartoon, art painting\} domain, urban examples are selected from the original cartoon domain, while non-urban examples are selected from the original art painting domain. In the \{art painting, cartoon\} domain, urban examples are selected from the original art painting domain, while non-urban examples are selected from the original cartoon domain. This sampling encourages the model to use spurious correlations (domain-related information) to predict the labels; however, since these relationships are flipped between domains \{\{cartoon,  art painting\} and \{art painting,  cartoon\}, these predictions will be wrong when generalized to other domains.

\paragraph{\em Terra Incognita:} The Terra Incognita dataset contains subsets of the Caltech Camera Traps dataset \citep{beery2018recognition} defined by \citep{Gul2020LostDG}. There are four domains representing different locations \{L100, L38, L43, L46\} of cameras in the American Southwest. There are 9 species of wild animals \{bird, bobcat, cat, coyote, dog, empty, opossum, rabbit, raccoon, squirrel\} and a `no-animal' class to be predicted. Like \cite{ahuja2021invariance}, we classify this dataset as following the generative process in Figure \ref{fig:third}, the Fully  Informative Invariant Features (FIIF) setting.
 Additional details on model architecture, training, and hyperparameters are detailed in Appendix \ref{sec:tcri_expr}.

\paragraph{Model Selection.} The standard approach for model selection is a training-domain hold-out validation set accuracy. We find that model selection across hyperparameters using this held-out training domain validation accuracy often returns non-domain-general models in the `hard' cases. One advantage of our model is that we can do model selection based on the TCRI condition (conditional independence between the two representations) on held-out training domain validation examples to mitigate this challenge. In the easy case, we expect the empirical risk minimizer to be domain-general, so selecting the best-performing training-domain model is sound -- we additionally do this for all baselines (see Appendix \ref{app:bench} for further discussion). We find that, empirically, this heuristic works in the examples we study in this work. Nevertheless, model selection under distribution shift remains a significant bottleneck for domain generalization.

\subsection{Results and Discussion}
\begin{table*}[h]
    \caption{$\Ecal \backslash e_{test} \rightarrow e_{test}$ (model selection on held-out source domains validation set). The `mean' column indicates the average generalization accuracy over all three domains as the $e_{test}$ distinctly; the `min' column indicates the worst generalization accuracy.}
    \centering
    \input{results/results}
    \label{tab:results}
\end{table*}

\begin{table*}[ht]
    \centering
    \caption{Total Information Criterion: Domain General (DG) and Domain Specific (DS) Accuracies. The DG classifier is shared across all training domains, and the DS classifiers are trained on each domain. The first row indicates the domain from which the held-out examples are sampled, and the second indicates which domain-specific predictor is used. \{+90\%, +80\%, -90\%\} indicate domains -- $\{0.1, 0.2, 0.9\}$ digit label and color correlation, respectively.} 
    \resizebox{\textwidth}{!}{
        \input{results/tic}
    }
    \label{tab:cmnist_tic}
\end{table*}

\paragraph{Worst-domain Accuracy.} A critical implication of domain generality is stability -- robustness in worst-domain performance up to domain difficulty. While average accuracy across domains provides some insight into an algorithm's ability to generalize to new domains, the average hides the variance of performance across domains. Average improvement can be increased while the worst-domain accuracy stays the same or decreases, leading to incorrect conclusions about domain generalization. Additionally, in real-world challenges such as algorithmic fairness where worst-group performance is considered, some metrics or fairness are analogous to achieving domain generalization~\citep{creager2021environment}.

\paragraph{Results.} TCRI achieves the highest average and worst-case accuracy across all baselines (Table \ref{tab:results}). We find no method recovers the exact domain-general model's accuracy of  $75\%$. However, TCRI achieves over 7\% increase in both average accuracy and worst-case accuracy. Appendix \ref{sec:app_cmnist}
shows transfer accuracies with cross-validation on held-out test domain examples (oracle) and TCRI again outperforms all baselines, achieving an average accuracy of 70.0\% $\pm$ 0.4\% and a worst-case accuracy of 65.7\% $\pm$ 1.5, showing that regularizing for TCRI gives very close to optimal domain-general solutions.

Similarly, for the Spurious-PACS dataset, we observe that TCRI outperforms the baselines. TRCI achieves the highest average accuracy of $63.4\% \pm 0.2$ and worst-case accuracy of $62.3\% \pm 0.1$ with the next best, VREx, achieving $58.8 \pm 1.0$ and $33.8 \pm 0.0$, respectively. Additionally, for the Terra-Incognita dataset, TCRI achieves the highest average and worst-case accuracies of 49.2\% $\pm$ 0.3\% and 40.4\% $\pm$ 1.6\% with the next best, GroupDRO, achieving $47.8 \pm 0.9$ and $39.9 \pm 0.7$, respectively.

Appendix \ref{sec:app_cmnist} shows transfer accuracies with cross-validation held-out target domain examples (oracle) where we observe that TCRI also obtains the highest average and worst-case accuracy for Spurrious-PACS and Terra Incognita.

Overall, regularizing for TCRI gives the most domain-general solutions compared to our baselines, achieving the highest worst-case accuracy on all benchmarks. Additionally, TCRI achieves the highest average accuracy on ColoredMNIST and Spurious-PAC and the second highest on Terra Incognita, where we expect the empirical risk minimizer to be domain-general.

Additional results are provided in the Appendix \ref{app:add_res}.

\begin{table}[t]
    \centering
    \caption{TIC ablation for ColoredMNIST.}

\input{results/tic_ablat}

    \label{tab:cmnist_tic_ablation}
\end{table}
\paragraph{The Effect of the Total Information Criterion.} Without the TIC loss term, our proposed method is less effective. Table \ref{tab:cmnist_tic_ablation}  shows that for Colored MNIST, the hardest `hard' case we encounter, removing the TIC criteria, performs worse in average and worst case accuracy, dropping over 8\% and 18\*, respectively.

\paragraph{Separation of Domain General and Domain Specific Features}. In the case of Colored MNIST, we can reason about the extent of feature disentanglement from the accuracies achieved by the domain-general and domain-specific predictors. Table \ref{tab:cmnist_tic} shows how much each component of $\Phi$, $\Phic$ and $\Phie$, behaves as expected. For each domain, we observe that the domain-specific predictors' accuracies follow the same trend as the color-label correlation, indicating that they capture the color-label relationship. The domain-general predictor, however, does not follow such a trend, indicating that it is not using color as the predictor. 

For example, when evaluating the domain-specific predictors from the +90\% test domain experiment (row +90\%) on held-out examples from the +80\% training domain (column "DS Classifier on +80\%"), we find that the +80\% domain-specific predictor achieves an accuracy of nearly 79.9\% -- exactly what one would expect from a predictor that uses a color correlation with the same direction `+'. Conversely, the -90\% predictor achieves an accuracy of 20.1\%, exactly what one would expect from a predictor that uses a color correlation with the opposite direction `-'. The -90\% domain has the opposite label-color pairing, so a color-based classifier will give the opposite label in any `+' domain.

Another advantage of this method, exemplified by Table \ref{tab:cmnist_tic}, is that if one believes a particular domain is close to one of the training domains, one can opt to use the close domain's domain-specific predictor and leverage spurious information to improve performance.

\paragraph{On Benchmarking Domain Generalization.} Previous work on benchmarking domain generalization showed that across standard benchmarks, the domain-unaware empirical risk minimizer outperforms or achieves equivalent performance to the state-of-the-art domain generalization methods \citep{Gul2020LostDG}. Additionally, \cite{Rosenfeld2021AnOL} gives results that show weak conditions that define regimes where the empirical risk minimizer across domains is optimal in both average and worst-case accuracy. Consequently, to accurately evaluate our work and baselines, we focus on settings where it is clear that (i) the empirical risk minimizer fails, (ii) spurious features, as we have defined them, do not generalize across the observed domains, and (iii) there is room for improvement via better domain-general predictions. We discuss this point further in the Appendix \ref{app:bench}.

\paragraph{Oracle Transfer Accuracies.} While model selection is an integral part of the machine learning development cycle, it remains a non-trivial challenge when there is a distribution shift. While we have proposed a selection process tailored to our method that can be generalized to other methods with an assumed causal graph, we acknowledge that model selection under distribution shift is still an important open problem. Consequently, we disentangle this challenge from the learning problem and evaluate an algorithm's capacity to give domain-general solutions independently of model selection. We report experimental reports using held-out test-set examples for model selection in Appendix \ref{app:add_res} Table \ref{tab:results_oracle}.  We find that our method, TCRI\_HSIC, also outperforms baselines in this setting.

%% file: results/simulated.tex
\begin{tabular}{c|c|c}
     \hline
     \textbf{Algorithm} & $\mathbf{(\Phic)_0}$ & $\mathbf{(\Phic)_1}$  \\ 
     & \textbf{(i.e.,} $\mathbf{\Zc}$ \textbf{weight)} & \textbf{(i.e., } $\mathbf{\Ze}$ \textbf{weight)}  \\ 
     \hline
     ERM            & 0.29 & 0.71 \\
     IRM            & 0.28 & 0.71 \\
     TCRI           & 1.01 & 0.06 \\
     \hline
     Oracle         & 1.04 & 0.00 \\
\end{tabular}

%% file: results/results.tex
\begin{tabular}{l?c|c?c|c|c|c}
\multicolumn{1}{c}{} &
\multicolumn{2}{c}{ColoredMNIST} &
\multicolumn{2}{c}{Spurious PACS} &
\multicolumn{2}{c}{Terra Incognita} \\
\hline
\textbf{Algorithm}      & \textbf{average}           & \textbf{worst-case}         & \textbf{average}           & \textbf{worst-case}         & \textbf{average}           & \textbf{worst-case}         \\
\hline
ERM                     & 51.6 $\pm$ 0.1             & 10.0 $\pm$ 0.1              & 57.2 $\pm$ 0.7              & 31.2 $\pm$ 1.3              & 44.2 $\pm$ 1.8              & 35.1 $\pm$ 2.8              \\  
IRM                     & 51.7 $\pm$ 0.1             & 9.9 $\pm$ 0.1               & 54.7 $\pm$ 0.8              & 30.3 $\pm$ 0.3              & 38.9 $\pm$ 3.7              & 32.6 $\pm$ 4.7              \\  
GroupDRO                & 52.0 $\pm$ 0.1             & 9.9 $\pm$ 0.1               & 58.5 $\pm$ 0.4              & 37.7 $\pm$ 0.7              & 47.8 $\pm$ 0.9              & 39.9 $\pm$ 0.7              \\  
VREx                    & 51.7 $\pm$ 0.2             & 10.2 $\pm$ 0.0              & 58.8 $\pm$ 0.4              & 37.5 $\pm$ 1.1              & 45.1 $\pm$ 0.4              & 38.1 $\pm$ 1.3              \\  
IB\_ERM                 & 51.5 $\pm$ 0.2             & 10.0 $\pm$ 0.1              & 56.3 $\pm$ 1.1              & 35.5 $\pm$ 0.4              & 46.0 $\pm$ 1.4              & 39.3 $\pm$ 1.1              \\  
IB\_IRM                 & 51.7 $\pm$ 0.0             & 9.9 $\pm$ 0.0               & 55.9 $\pm$ 1.2              & 33.8 $\pm$ 2.2              & 37.0 $\pm$ 2.8              & 29.6 $\pm$ 4.1              \\   
\hline
TCRI\_HSIC              & \textbf{59.6 $\pm$ 1.8}    & \textbf{45.1 $\pm$ 6.7}     & \textbf{63.4 $\pm$ 0.2}     & \textbf{62.3 $\pm$ 0.2}     & \textbf{49.2 $\pm$ 0.3}     & \textbf{40.4 $\pm$ 1.6}    \\
\end{tabular}

%% file: results/tic.tex
\begin{tabular}{l?ccc?ccc|ccc|ccc}
{} & \multicolumn{3}{c?}{\textbf{DG Classifier}} & \multicolumn{3}{c}{\textbf{DS Classifier on +90 }} & \multicolumn{3}{c}{\textbf{DS Classifier on +80}} & \multicolumn{3}{c}{\textbf{DS Classifier on -90}}\\
\hline
\vtop{\hbox{\strut \textbf{Test Domain}}\hbox{\strut No DS clf.}} &   +90\% &  +80\% &  -90\% &   +90\% &  +80\% &  -90\% &  +90\% & +80\% &  -90\% & +90\% & +80\% &  -90\% \\
\hline
+90\%        &          68.7 &          69.0 &          68.5 &                     - &                    90.1 &                     9.8 &                     - &                    \cellcolor{yellow!50}79.9 &                    \cellcolor{yellow!50}20.1 &                     - &                    10.4 &                    89.9 \\
+80\%        &          63.1 &          62.4 &          64.4 &                    76.3 &                     - &                    24.3 &                    70.0 &                     - &                    30.4 &                    24.5 &                     - &                    76.3 \\
-90\%        &          65.6 &          63.4 &          44.1 &                    75.3 &                    75.3 &                     - &                    69.2 &                    69.5 &                     - &                    29.3 &                    26.0 &                     - \\
\end{tabular}

%% file: results/tic_ablat.tex
\begin{tabular}{l|c|c}
     \textbf{Algorithm} & \textbf{average} & \textbf{worst-case} \\
     \hline
     TCRI\_HSIC (No TIC) & 51.8 $\pm$ 5.9 & 27.7 $\pm$ 8.9 \\
     TCRI\_HSIC          & {\bf 59.6 $\pm$ 1.8} & {\bf 45.1 $\pm$ 6.7}
\end{tabular}

%% file: 00_paper/07_Conclusion.tex
\section{Conclusion and Future Work}
We reduce the gap in learning domain-general predictors by leveraging conditional independence properties implied by generative processes to identify domain-general mechanisms. We do this without independent observations of domain-general and spurious mechanisms and show that our framework outperforms other state-of-the-art domain-generalization algorithms on real-world datasets in average and worst-case across domains. Future work includes further improvements to the framework to fully recover the strict set of domain-general mechanisms and model selection strategies that preserve desired domain-general properties.

\section*{Acknowledgements}
OS was partially supported by the UIUC Beckman Institute Graduate Research Fellowship, NSF-NRT 1735252. This work is partially supported by the NSF III 2046795, IIS 1909577, CCF 1934986, NIH 1R01MH116226-01A, NIFA award 2020-67021-32799, the Alfred P. Sloan Foundation, and Google Inc.

%% file: 00_paper/99_appendix_expr.tex
\onecolumn
\section{Additional Results and Discussion} \label{app:add_res}
\subsection{On Benchmarking Domain Generalization} \label{app:bench}
\begin{table*}[h]
    \caption{Oracle (model selection on held-out target domain validation set) $\Ecal \backslash e_{test} \rightarrow e_{test}$. The `mean' column indicates the average generalization accuracy over all three domains as the $e_{test}$ distinctly; the `min' column indicates the worst generalization accuracy.}
    \centering
    \input{results/cmnist_results_oracle}
    \label{tab:results_oracle}
\end{table*}

\paragraph{Oracle Transfer Accuracies.} While model selection is an integral part of the machine learning development cycle, it remains a non-trivial challenge when there is a distribution shift. While we have proposed a selection process tailored to our method that can be generalized to other methods with an assumed causal graph, we acknowledge that model selection under distribution shift is still an important open problem. Consequently, we disentangle this challenge from the learning problem and evaluate an algorithm's capacity to give domain-general solutions independently of model selection. We report experimental reports using held-out test-set examples for model selection in Appendix \ref{app:add_res} Table \ref{tab:results_oracle}. 

In this case, we find that there is indeed a separation between ERM and some domain-generalization algorithms, suggesting that model selection might be a substantial bottleneck for learning domain-general predictors. Nevertheless, we still find that our method, TCRI\_HSIC, also outperforms baselines in this setting.

\paragraph{Challenges of Benchmarking Domaing Generalization.} We show some results below that illustrate the challenge of accurately evaluating the efficacy of an algorithm for domain generalization. We first note that we expect ERM (naive) to perform poorly in domain generalization tasks, certainly so when we observe worst-case shifts at test time. However, like other works \citep{Gul2020LostDG}, we observe that ERM performs as well as other baselines during transfer on various benchmark datasets. Previous theoretical results \citep{Rosenfeld2021AnOL} suggest that this observation is indicative of properties of the benchmark domains that may be sufficient for ERM to give domain-general solutions - specifically that the distribution (and equivalently the loss) of the target domain can be written as a convex combination of the those in the source domains.

To further investigate this, we develop additional experiments motivated by the ColoredMNIST \citep{Arjovsky2019InvariantRM} -- since its generative process is well understood. We note that in the +90\%, +80\%, and -90\% domains of ColoredMNIST, the -90\% domain has the opposite relationship between the spurious correlation and the label, so the use of spurious correlations from \{+90\%, +80\%\} generalizes catastrophically to the -90\% domain. In this setting, the baseline algorithms we present, including ERM, achieve poor accuracy in the -90\% domain while maintaining high accuracy in the +90\% and +80\% domains. Consequently, we investigate two settings, \emph{setting a}: observe \{+90\%, +80\%, +70\%, -90\%\} domains and \emph{setting b}: observe \{+90\%, +80\%, -80\%, -90\%\} domains -- we focus on generalizing to the -90\% domain. In {\em setting a}, we add another domain with the majority direction in the relationship between spurious correlation and labels. In {\em setting b}, we add another domain with the minority direction. Note that in {\em setting a}, the closest domain to -90\% that can be generated with a convex combination of the other domains still has a `+' correlation between the color and label. In {\em setting b}, however, one can generate a domain with a `-' correlation between color and label with a convex combination of the other domains. Thus, we expect the empirical risk minimizer to give domain-general solutions in setting b but not in setting a.

We use Oracle model selection (held-out target data) to remove the effect of model selection for all methods in the results. We find that in setting a, where we add a domain (+70\%), we observe that the generalization accuracy to the -90\% domain is still very different from the other domains (Table \ref{tab:cmnist3_oracle}).

\begin{table*}[h]
    \centering
    \renewcommand{\arraystretch}{1.1}
    \caption{ColoredMNIST \emph{setting a}. Columns \{+90\%, +80\%, +70\%, -90\%\} indicate domains -- $\{0.1, 0.2, 0.3, 0.9\}$ digit label and color correlation, respectively. We report domain accuracies over 3 trials each. We use the oracle selection method -- held out target data. $\Ecal \backslash e_{test} \rightarrow e_{test}$.}
    \input{results/cmnist3_oracle}
    \label{tab:cmnist3_oracle}
\end{table*}

However, in setting b, where we add a domain (-80\%), we observe that the generalization accuracy to the -90\% domain is on par with the other domains (Table \ref{tab:cmnist8_oracle}).
\begin{table*}[h]
    \centering
    \renewcommand{\arraystretch}{1.1}
    \caption{ColoredMNIST \emph{setting b}. Columns \{+90\%, +80\%, -80\%, -90\%\} indicate domains -- $\{0.1, 0.2, 0.8, 0.9\}$ digit label and color correlation, respectively. We report the average domain accuracies over 3 trials each. We use the oracle selection method -- held out target data. $\Ecal \backslash e_{test} \rightarrow e_{test}$.}
    \input{results/cmnist8_oracle}
    \label{tab:cmnist8_oracle}
\end{table*}

This illustrates the challenge of accurately evaluating an algorithm's ability to give domain-general predictions. We note that it is generally difficult to distinguish between {\em setting a} and {\em setting b}. The primary signature we see is some consistency between the empirical risk minimizer and the other baselines. \cite{Gul2020LostDG} observe a similar trend for standard benchmarks for domain generalization. Hence, we focus our empirical evaluations in this work on settings where we know that the ERM solution fails by design.

\subsection{ColoredMNIST} \label{sec:app_cmnist}
{\em ColoredMNIST:} The ColoredMNIST dataset \citep{Arjovsky2019InvariantRM} is composed of $7000$ ($2\times 28 \times 28$, $1$) images of a hand-written digit and binary-label pairs. There are three domains with different correlations between image color and label, i.e., the image color is spuriously related to the label by assigning a color to each of the two classes (0: digits 0-4, 1: digits 5-9). The color is then flipped with probabilities $\{0.1, 0.2, 0.9\}$ to create three domains, making the color-label relationship domain-specific because it changes across domains. There is also label flip noise of $0.25$, so we expect that the best accuracy a domain-general model can achieve is 75\%, while a non-domain general model can achieve higher. In this dataset, $\Zc$ corresponds to the original image, $\Ze$ the color, $e$ the label-color correlation, $Y$ the image label, and $X$ the observed colored image. This DAG follows the generative process of Figure \ref{fig:first} 

\begin{table}[h]
    \centering
    \caption{ColoredMNIST Hyperparameters. Additional hyperparameters are provided in \href{https://github.com/olawalesalaudeen/tcri}{https://github.com/olawalesalaudeen/tcri}.}
    \begin{tabular}{c|c|c|c}
         Algorithm & Hyperparameter & Default & Random Distribution\\
         \hline
         \multirow{2}{*}{All} & Learning Rate & $1^{-3}$ & $10^{\text{Uniform}(-4.5, -2.5)}$ \\
         & Batch Size & 64 & $2^{\text{Uniform}(3, 9)}$ \\
         \hline
         \multirow{2}{*}{TCRI $\beta$} & penalty weight & $100$ & $10^\text{Uniform}(-1, 5)$ \\
         & annealing steps & $500$ & $10^\text{Uniform}(2.5, 5)$ \\
         
    \end{tabular}
    \label{apptab:cmnist_hparams}
\end{table}

\begin{table}[h]
    \centering
    \caption{MNIST ConvNet architecture. All convolutions use 3$\times$3 kernels and "same" padding.}
    \begin{tabular}{c|c}
    \textbf{\#} & \textbf{Layer} \\
    \hline
        1 & Conv2D (in=d, out=64) \\
        2 & ReLU \\
        3 & GroupNorm (groups=8) \\
        4 & Conv2D (in=64, out=128, stride=2) \\
        5 & ReLU \\
        6 & GroupNorm (groups=8) \\
        7 & Conv2D (in=128, out=128) \\
        8 & ReLU \\
        9 & GroupNorm (groups=8) \\
        10 & Conv2D (in=128, out=128) \\
        11 & ReLU \\
        12 & GroupNorm (8 groups) \\
        13 & Global average-pooling \\
    \end{tabular}
    \label{apptab:mnist_convnet}
\end{table}

We use MNIST-ConvNet \citep{Gul2020LostDG} backbones for the MNIST datasets (Table \ref{apptab:mnist_convnet}). Both $\Phic$ and $\Phie$ are linear layers of size $128 \times 128$ that are appended to the backbone. The predictors (classification hyperplanes) $\theta_c,\, \{\theta_1,\,\theta_{2}\}$ are also parameterized to be linear and appended to the $\Phic$ and $\Phi$, respectively.

We do a random search to select hyperparameters using the same scheme as \cite{Gul2020LostDG} (\href{https://github.com/facebookresearch/DomainBed}{https://github.com/facebookresearch/DomainBed}). We select 25 hyperparameters with 5 random restarts each to generate error bars. 

We show transfer accuracies with both source and target domain validation for model selection in Tables \ref{apptab:cmnist_env}-\ref{apptab:cmnist_oracle}. We find that TCRI outperforms all baselines in average and worst-case accuracy.

\begin{table}[h]
    \centering
    \caption{ColoredMNIST Transfer Accuracy -- model selection on held-out source validation set. Columns \{+90\%, +80\%, -90\%\} indicate domains -- $\{0.1, 0.2, 0.9\}$ digit label and color correlation, respectively. $\Ecal \backslash e_{test} \rightarrow e_{test}$.}
    \label{apptab:cmnist_env}
    \input{results/cmnist_results_env}
\end{table}

\begin{table}[!h]
    \centering
    \caption{Oracle ColoredMNIST Transfer Accuracy -- model selection on held-out target validation set accuracy. Columns \{+90\%, +80\%, -90\%\} indicate domains -- $\{0.1, 0.2, 0.9\}$ digit label and color correlation, respectively. $\Ecal \backslash e_{test} \rightarrow e_{test}$.}
    \label{apptab:cmnist_oracle}
    \input{results/cmnist_results_oracle}
\end{table}

\subsection{Spurrious PACS}
{\em Spurious--PACS.} \emph{Variables.} $X$: images, $Y$: non-urban (elephant, giraffe, horse) vs. urban (dog, guitar, house, person). \emph{Domains.} \{\{cartoon,  art painting\}, \{art painting,  cartoon\}, \{photo\}\} \citep{li2017deeper}. The photo domain is the same as in the original dataset. In the \{cartoon, art painting\} domain, urban examples are selected from the original cartoon domain, while non-urban examples are selected from the original art painting domain. In the \{art painting, cartoon\} domain, urban examples are selected from the original art painting domain, while non-urban examples are selected from the original cartoon domain. This sampling encourages the model to use spurious correlations (domain-related information) to predict the labels; however, since these relationships are flipped between domains \{\{cartoon,  art painting\} and \{art painting,  cartoon\}, these predictions will be wrong when generalized to other domains.

\begin{table}[h]
    \centering
    \caption{Spurrious PACS Hyperparameters.  Additional hyperparameters provided in \href{https://github.com/olawalesalaudeen/tcri}{https://github.com/olawalesalaudeen/tcri}.}
    \begin{tabular}{c|c|c|c}
         Algorithm & Hyperparameter & Default & Range \\
         \hline
         \multirow{2}{*}{All} & Learning Rate & $1^{-3}$ & $10^{\text{Uniform}(-4.5, -2.5)}$ \\
         & Batch Size & 64 & $2^{\text{Uniform}(3, 9)}$ \\
         \hline
         \multirow{2}{*}{TCRI $\beta$} & penalty weight & $100$ & $10^\text{Uniform}(-1, 5)$ \\
         & annealing steps & $500$ & $10^\text{Uniform}(2.5, 5)$
    \end{tabular}
    \label{apptab:pacs_v3_hparams}
\end{table}

We use a ResNet-50 backbone \citep{he2016deep}. $\Phic$ and $\Phie$ are linear layers of size $2048 \times 2048$ that are appended to the backbone. The predictors (classification hyperplanes) $\theta_c, \{\theta_1, \theta_2, \theta_3\}$ are linear and appended to $\Phic$ and $\Phi$ layers, respectively.

\paragraph{Hyperparameters:} We do a random search to select hyperparameters using the same scheme as \cite{Gul2020LostDG} (\href{https://github.com/facebookresearch/DomainBed}{https://github.com/facebookresearch/DomainBed}). We select 5 hyperparameters with 3 random restarts each to generate error bars. 

We show transfer accuracies with both source and target domain validation for model selection in Tables \ref{apptab:pacs_v3}-\ref{apptab:oracle_pacs_v3}. We find that TCRI outperforms all baselines in average and worst-case accuracy.

\begin{table*}[!h]
    \centering
    \caption{Spurious--PACS Transfer Accuracy -- model selection on held-out source validation set. $\Ecal \backslash e_{test} \rightarrow e_{test}$.}
    \label{apptab:pacs_v3}
    \renewcommand{\arraystretch}{1.1}
    \input{results/pacs_v3}
\end{table*}

\begin{table*}[!h]
    \centering
    \caption{Oracle Spurious--PACS Transfer Accuracy -- model selection on held-out target validation set. $\Ecal \backslash e_{test} \rightarrow e_{test}$.}
    \label{apptab:oracle_pacs_v3}
    \renewcommand{\arraystretch}{1.1}
    \input{results/pacs_v3_oracle}
\end{table*}

\FloatBarrier
\subsection{Terra Incognita}
The Terra Incognita dataset contains subsets of the Caltech Camera Traps dataset \citep{beery2018recognition} defined by \citep{Gul2020LostDG}. Four domains represent different locations \{L100, L38, L43, L46\} of cameras in the American Southwest. There are 10 different species of wild animals \{bird, bobcat, cat, coyote, dog, empty, opossum, rabbit, raccoon, squirrel\} (classes) to be predicted. Like \cite{ahuja2021invariance}, we classify this dataset as following the generative process in Figure \ref{fig:third}, the Fully  Informative Invariant Features (FIIF) setting.

\begin{table}[h]
    \centering
    \caption{Terra Incognita Hyperparameters. Additional hyperparameters provided in \href{https://github.com/olawalesalaudeen/tcri}{https://github.com/olawalesalaudeen/tcri}.}
    \begin{tabular}{c|c|c|c}
         Algorithm & Hyperparameter & Default & Range \\
         \hline
         \multirow{2}{*}{All} & Learning Rate & $1^{-3}$ & $10^{\text{Uniform}(-4.5, -2.5)}$ \\
         & Batch Size & 64 & $2^{\text{Uniform}(3, 9)}$ \\
         \hline
         \multirow{2}{*}{TCRI $\beta$} & penalty weight & $100$ & $10^\text{Uniform}(-1, 5)$ \\
         & annealing steps & $500$ & $10^\text{Uniform}(0, 4)$
    \end{tabular}
    \label{apptab:terra_hparams}
\end{table}

We use a ResNet-50 backbone \citep{he2016deep}. $\Phic$ and $\Phie$ are linear layers of size $2048 \times 2048$ that are appended to the backbone. The predictors (classification hyperplanes) $\theta_c, \{\theta_1, \theta_2, \theta_3, \theta_4\}$ are linear and appended to $\Phic$ and $\Phi$ layers, respectively.

\paragraph{Hyperparameters:} We do a random search to select hyperparameters using the same scheme as \cite{Gul2020LostDG} (\href{https://github.com/facebookresearch/DomainBed}{https://github.com/facebookresearch/DomainBed}). We select 5 hyperparameters with 3 random restarts each to generate error bars.

We show transfer accuracies with both source and target domain validation for model selection in Tables \ref{apptab:terra_env}-\ref{apptab:terra_oracle}. We find that TCRI outperforms all baselines except ERM on average and outperforms all baselines in worst-case accuracy.

\begin{table}[h]
    \centering
    \caption{Terra Incognita Transfer Accuracy -- model selection on held-out source validation set. $\Ecal \backslash e_{test} \rightarrow e_{test}$.}
    \label{apptab:terra_env}
    \resizebox{\textwidth}{!}{
    \input{results/terra_results_env}
    }
\end{table}

\begin{table}[h]
    \centering
    \caption{Oracle Terra Incognita Transfer Accuracy -- model selection on held-out target validation set. $\Ecal \backslash e_{test} \rightarrow e_{test}$.}
    \label{apptab:terra_oracle}
    \resizebox{\textwidth}{!}{
    \input{results/terra_results_env_oracle}
    }
\end{table}

%% file: results/cmnist_results_oracle.tex
\begin{tabular}{l?c|c?c|c|c|c}
\multicolumn{1}{c}{} & \multicolumn{2}{c}{ColoredMNIST} & \multicolumn{2}{c}{Spurious PACS} & \multicolumn{2}{c}{Terra Incognita} \\
\hline
\textbf{Algorithm} & \textbf{average} & \textbf{worst-case} & \textbf{average} & \textbf{worst-case} & \textbf{average} & \textbf{worst-case} \\
\hline
ERM                & 57.8 $\pm$ 0.2            & 38.4 $\pm$ 1.4            & 59.2 $\pm$ 1.3                  & 38.4 $\pm$ 1.4     & {\bf 52.9 $\pm$ 0.8} & 42.0 $\pm$ 0.6           \\
IRM                & 68.9 $\pm$ 1.6            & 62.0 $\pm$ 4.9            & 67.5 $\pm$ 5.8                  & 53.9 $\pm$ 6.6     & 42.6 $\pm$ 4.0        & 42.7 $\pm$ 1.2           \\
GroupDRO           & 61.1 $\pm$ 1.3            & 37.6 $\pm$ 3.6            & 61.8 $\pm$ 1.8                  & 40.0 $\pm$ 1.6     & 50.7 $\pm$ 1.0        & 42.7 $\pm$ 1.2           \\
VREx               & 68.0 $\pm$ 2.5            & 59.4 $\pm$ 7.3            & 62.8 $\pm$ 2.4                  & 38.7 $\pm$ 0.9     & 43.2 $\pm$ 2.0        & 34.9 $\pm$ 4.2           \\
IB\_ERM            & 65.0 $\pm$ 0.1            & 50.6 $\pm$ 0.3            & 67.3 $\pm$ 3.7                  & 53.1 $\pm$ 8.0     & 49.0 $\pm$ 0.3        & 39.9 $\pm$ 0.8           \\
IB\_IRM            & 68.4 $\pm$ 1.0            & 58.5 $\pm$ 2.8            & 69.0 $\pm$ 1.3                  &{\bf 62.3 $\pm$ 0.3}& 32.8 $\pm$ 6.6        & 20.4 $\pm$ 7.5           \\
\hline
TCRI\_HSIC         & {\bf 70.4 $\pm$ 0.4}      & {\bf 65.7 $\pm$ 1.5}      & {\bf 69.5 $\pm$ 1.1}           & {\bf 62.3 $\pm$ 0.2}& 51.2 $\pm$ 0.1       & {\bf 43.0 $\pm$ 0.4}     \\
\end{tabular}

%% file: results/cmnist3_oracle.tex
\begin{tabular}{l|c|c|c|c} 
\textbf{Algorithm}   & \textbf{+90\%}       & \textbf{+80\%}       & \textbf{+70\%}       & \textbf{-90\%}       \\ 
\hline
ERM                  & 72.8 $\pm$ 0.3       & 74.7 $\pm$ 0.3       & 73.3 $\pm$ 0.1       & 16.3 $\pm$ 1.5      \\ 
IRM                  & 49.0 $\pm$ 0.1       & 54.2 $\pm$ 2.0       & 50.3 $\pm$ 0.3       & 43.8 $\pm$ 2.8      \\ 
GroupDRO             & 71.0 $\pm$ 0.6       & 72.2 $\pm$ 0.3       & 70.7 $\pm$ 0.9       & 36.4 $\pm$ 4.2      \\ 
VREx                 & 74.1 $\pm$ 1.3       & 72.6 $\pm$ 0.5       & 72.1 $\pm$ 0.5       & 19.5 $\pm$ 5.5      \\ 

\hline 
TCRI\_HSIC             & 72.1 $\pm$ 1.5       & 73.6 $\pm$ 0.4       & 72.6 $\pm$ 0.4       & 49.9 $\pm$ 0.3      \\ 
\end{tabular}

%% file: results/cmnist8_oracle.tex
\begin{tabular}{l|c|c|c|c} 
\textbf{Algorithm}   & \textbf{+90\%}       & \textbf{+80\%}       & \textbf{-80\%}       & \textbf{-90\%}       \\ 
\hline
ERM                  & 58.4 $\pm$ 1.3       & 67.0 $\pm$ 0.5       & 64.2 $\pm$ 2.0       & 52.6 $\pm$ 3.2      \\ 
IRM                  & 56.7 $\pm$ 3.3       & 56.6 $\pm$ 2.8       & 51.6 $\pm$ 0.7       & 51.7 $\pm$ 0.7      \\ 
GroupDRO             & 69.7 $\pm$ 0.8       & 71.7 $\pm$ 0.3       & 72.0 $\pm$ 0.2       & 71.4 $\pm$ 1.9      \\ 
VREx                 & 67.4 $\pm$ 1.9       & 70.4 $\pm$ 0.1       & 71.2 $\pm$ 0.2       & 59.4 $\pm$ 4.3      \\ 
\hline
TCRI\_HSIC           & 62.2 $\pm$ 4.4       & 70.0 $\pm$ 1.3       & 67.9 $\pm$ 1.4       & 65.4 $\pm$ 2.8      \\ 
\end{tabular}

%% file: results/cmnist_results_env.tex
\begin{tabular}{l?ccc?ccc}
 &
 \multicolumn{3}{c?}{Domains} &
 \multicolumn{3}{c}{Domain Accuracy Statistics}\\
 \hline
\textbf{Algorithm}   & \textbf{+90\%}       & \textbf{+80\%}       & \textbf{-90\%}       & \textbf{Avg}         & \textbf{Std}         & \textbf{Min}         \\
\hline
ERM                  & 71.6 $\pm$ 0.3       & 73.1 $\pm$ 0.1       & 10.0 $\pm$ 0.1       & 51.6 $\pm$ 0.1       & 29.4 $\pm$ 0.1       & 10.0 $\pm$ 0.1       \\  
IRM                  & 72.1 $\pm$ 0.1       & 73.0 $\pm$ 0.3       & 9.9 $\pm$ 0.1        & 51.7 $\pm$ 0.1       & 29.5 $\pm$ 0.1       & 9.9 $\pm$ 0.1        \\  
GroupDRO             & 72.6 $\pm$ 0.2       & 73.4 $\pm$ 0.2       & 9.9 $\pm$ 0.1        & 52.0 $\pm$ 0.1       & 29.8 $\pm$ 0.1       & 9.9 $\pm$ 0.1        \\  
VREx                 & 72.2 $\pm$ 0.2       & 72.7 $\pm$ 0.3       & 10.2 $\pm$ 0.0       & 51.7 $\pm$ 0.2       & 29.3 $\pm$ 0.1       & 10.2 $\pm$ 0.0       \\  
IB\_ERM              & 71.0 $\pm$ 0.4       & 73.4 $\pm$ 0.3       & 10.0 $\pm$ 0.1       & 51.5 $\pm$ 0.2       & 29.4 $\pm$ 0.1       & 10.0 $\pm$ 0.1       \\  
IB\_IRM              & 71.7 $\pm$ 0.2       & 73.4 $\pm$ 0.1       & 9.9 $\pm$ 0.0        & 51.7 $\pm$ 0.0       & 29.5 $\pm$ 0.0       & 9.9 $\pm$ 0.0        \\  
\hline
TCRI\_HSIC           & 67.2 $\pm$ 2.3       & 65.6 $\pm$ 3.4       & 45.9 $\pm$ 6.9       & {\bf 59.6 $\pm$ 1.8}       & {\bf 11.4 $\pm$ 3.3}       & {\bf 45.1 $\pm$ 6.7}       \\
\end{tabular}

%% file: results/pacs_v3.tex
\begin{tabular}{l|c|c|c|ccc}
&
\multicolumn{3}{c!}{Domains} &
\multicolumn{3}{c!}{Domain Accuracy Statistics}\\
\hline
\textbf{Algorithm} & \textbf{C x A} & \textbf{A x C} & \textbf{P} & \textbf{mean} & \textbf{std} & \textbf{min}\\
\hline
ERM                  & 31.2 $\pm$ 1.3       & 42.8 $\pm$ 0.7       & 97.6 $\pm$ 0.2       & 57.2 $\pm$ 0.7                 & 29.0 $\pm$ 0.4                 & 31.2 $\pm$ 1.3                 \\
IRM                  & 30.3 $\pm$ 0.3       & 39.0 $\pm$ 1.3       & 94.9 $\pm$ 1.4       & 54.7 $\pm$ 0.8                 & 28.6 $\pm$ 0.8                 & 30.3 $\pm$ 0.3                 \\
GroupDRO             & 37.7 $\pm$ 0.7       & 42.1 $\pm$ 1.6       & 95.7 $\pm$ 0.5       & 58.5 $\pm$ 0.4                 & 26.4 $\pm$ 0.3                 & 37.7 $\pm$ 0.                 \\
VREx                 & 37.5 $\pm$ 1.1       & 43.0 $\pm$ 0.5       & 95.7 $\pm$ 1.5       & 58.8 $\pm$ 0.4                & 26.2 $\pm$ 1.0                 & 37.5 $\pm$ 1.1                 \\
IB\_{ERM}            & 35.5 $\pm$ 0.4       & 48.6 $\pm$ 3.3       & 84.8 $\pm$ 0.6       & 56.3 $\pm$ 1.1               & 20.8 $\pm$ 0.6                & 35.5 $\pm$ 0.4                 \\
IB\_{IRM}            & 33.8 $\pm$ 2.2       & 38.8 $\pm$ 3.0       & 95.1 $\pm$ 1.5       & 55.9 $\pm$ 1.2                 & 27.8 $\pm$ 1.5                 & 33.8 $\pm$ 0.4                 \\
\hline
TCRI\_HSIC       & 62.8 $\pm$ 0.1       & 62.3 $\pm$ 0.2       & 65.0 $\pm$ 0.4       & \textbf{63.4 $\pm$ 0.2}                 & \textbf{1.2$ \pm$ 0.2}                  & \textbf{62.3 $\pm$ 0.2}                 \\
\end{tabular}

%% file: results/pacs_v3_oracle.tex
\begin{tabular}{l|c|c|c?c|c|c}
\multicolumn{3}{c?}{Domains} & \multicolumn{3}{c}{Domain Accuracy Statistics}\\
\hline
\textbf{Algorithm} & \textbf{C x A} & \textbf{A x C} & \textbf{P} & \textbf{mean} & \textbf{std} & \textbf{min}\\
\hline
ERM                  & 38.4 $\pm$ 1.4       & 43.4 $\pm$ 1.9       & 95.9 $\pm$ 0.6      & 59.2                 & 26.0                 & 38.4                 \\
IRM                  & 62.8 $\pm$ 0.1       & 53.9 $\pm$ 6.6       & 85.8 $\pm$ 8.2      & 67.5                 & 13.4                 & 53.9                 \\
GroupDRO             & 40.0 $\pm$ 1.6       & 49.7 $\pm$ 2.9       & 95.7 $\pm$ 0.6      & 61.8                 & 24.3                 & 40.0                 \\
VREx                 & 55.8 $\pm$ 5.5       & 38.7 $\pm$ 0.9       & 93.8 $\pm$ 0.8      & 62.8                 & 23.0                 & 38.7                 \\
IB\_{ERM}             & 53.1 $\pm$ 8.0       & 55.4 $\pm$ 5.7       & 93.5 $\pm$ 1.8      & 67.3                 & 18.5                 & 53.1                 \\
IB\_{IRM}             & 62.8 $\pm$ 0.1       & 62.3 $\pm$ 0.3       & 81.8 $\pm$ 7.0      & 69.0                 & {\bf 9.1}                  & {\bf 62.3}                 \\
\hline 
TCRI\_HSIC            & 64.0 $\pm$ 0.7       & 62.3 $\pm$ 0.2       & 82.4 $\pm$ 5.7      & {\bf 69.5}                 & {\bf 9.1}                  & {\bf 62.3}                 \\
\end{tabular}

%% file: results/terra_results_env.tex
\begin{tabular}{l?cccc?ccc}
 &
 \multicolumn{4}{c?}{Domains} &
 \multicolumn{3}{c}{Domain Accuracy Statistics}\\
 \hline
\textbf{Algorithm}   & \textbf{L100}        & \textbf{L38}         & \textbf{L43}         & \textbf{L46}         & \textbf{Avg}         & \textbf{Std}         & \textbf{Min}         \\
\hline
ERM                  & 43.6 $\pm$ 3.9       & 45.2 $\pm$ 0.6       & 53.0 $\pm$ 1.2       & 35.1 $\pm$ 2.8       & 44.2 $\pm$ 1.8       & 6.8 $\pm$ 1.0        & 35.1 $\pm$ 2.8       \\  
IRM                  & 43.9 $\pm$ 3.3       & 35.7 $\pm$ 4.0       & 37.7 $\pm$ 7.8       & 38.3 $\pm$ 2.4       & 38.9 $\pm$ 3.7       & {\bf 5.4 $\pm$ 1.8}        & 32.6 $\pm$ 4.7       \\  
GroupDRO             & 53.8 $\pm$ 4.6       & 40.5 $\pm$ 0.7       & 55.3 $\pm$ 1.5       & 41.8 $\pm$ 1.1       & 47.8 $\pm$ 0.9       & 7.7 $\pm$ 0.9        & 39.9 $\pm$ 0.7       \\  
VREx                 & 48.8 $\pm$ 2.0       & 38.1 $\pm$ 1.3       & 54.4 $\pm$ 0.6       & 39.0 $\pm$ 1.4       & 45.1 $\pm$ 0.4       & 7.0 $\pm$ 0.9        & 38.1 $\pm$ 1.3       \\  
IB\_ERM              & 46.1 $\pm$ 4.5       & 40.7 $\pm$ 0.7       & 55.2 $\pm$ 0.8       & 42.2 $\pm$ 1.1       & 46.0 $\pm$ 1.4       & 6.4 $\pm$ 0.8        & 39.3 $\pm$ 1.1       \\  
IB\_IRM              & 39.7 $\pm$ 7.3       & 40.8 $\pm$ 2.3       & 34.7 $\pm$ 4.3       & 32.9 $\pm$ 2.6       & 37.0 $\pm$ 2.8       & 6.7 $\pm$ 1.3        & 29.6 $\pm$ 4.1       \\ 
\hline
TCRI\_HSIC           & 54.6 $\pm$ 2.4       & 48.6 $\pm$ 2.0       & 53.2 $\pm$ 1.0       & 40.4 $\pm$ 1.6       & {\bf 49.2 $\pm$ 0.3}       & 6.1 $\pm$ 1.1        & {\bf 40.4 $\pm$ 1.6}       \\
\end{tabular}

%% file: results/terra_results_env_oracle.tex
\begin{tabular}{l?cccc?ccc}
 &
 \multicolumn{4}{c?}{Domains} &
 \multicolumn{3}{c}{Domain Accuracy Statistics}\\
 \hline
\textbf{Algorithm}   & \textbf{L100}        & \textbf{L38}         & \textbf{L43}         & \textbf{L46}         & \textbf{Avg}         & \textbf{Std}         & \textbf{Min}         \\
\hline
ERM                  & 58.5 $\pm$ 1.8       & 52.0 $\pm$ 1.3       & 59.2 $\pm$ 0.2       & 42.0 $\pm$ 0.6       & {\bf 52.9 $\pm$ 0.8}       & 7.0 $\pm$ 0.5        & 42.0 $\pm$ 0.6       \\
IRM                  & 53.0 $\pm$ 0.9       & 48.0 $\pm$ 1.8       & 36.3 $\pm$ 9.6       & 33.2 $\pm$ 3.9       & 42.6 $\pm$ 4.0       & 9.6 $\pm$ 1.7        & 30.8 $\pm$ 5.4       \\
GroupDRO             & 56.2 $\pm$ 3.0       & 45.2 $\pm$ 2.3       & 58.0 $\pm$ 0.2       & 43.3 $\pm$ 0.7       & 50.7 $\pm$ 1.0       & 6.9 $\pm$ 0.9        & 42.7 $\pm$ 1.2       \\
VREx                 & 43.2 $\pm$ 1.5       & 49.3 $\pm$ 1.2       & 41.5 $\pm$ 7.8       & 38.9 $\pm$ 1.1       & 43.2 $\pm$ 2.0       & 6.5 $\pm$ 1.8        & 34.9 $\pm$ 4.2       \\
IB\_ERM              & 55.6 $\pm$ 1.7       & 47.2 $\pm$ 1.1       & 53.4 $\pm$ 0.7       & 39.9 $\pm$ 0.8       & 49.0 $\pm$ 0.3       & 6.4 $\pm$ 0.5        & 39.9 $\pm$ 0.8       \\
IB\_IRM              & 40.2 $\pm$ 8.2       & 31.9 $\pm$ 11.8      & 29.4 $\pm$ 4.4       & 29.7 $\pm$ 3.8       & 32.8 $\pm$ 6.6       & 8.2 $\pm$ 1.0        & 20.4 $\pm$ 7.5       \\ 
\hline
TCRI\_HSIC           & 57.7 $\pm$ 1.8       & 50.1 $\pm$ 1.8       & 54.1 $\pm$ 0.6       & 43.0 $\pm$ 0.4       & 51.2 $\pm$ 0.1       & {\bf 5.8 $\pm$ 0.7}        & {\bf 43.0 $\pm$ 0.4}       \\
\end{tabular}

%% file: 00_paper/99_appendix_theory.tex
\section{DAGs} \label{sec:app_dags}

\FloatBarrier
\begin{figure}[h]
    \centering
    \input{files/graph_no_ey_x}
    \caption{Partial Ancestral Graph (PAG). Dashed edges indicate that the edge may or may not exist. The combination of $Y \rightarrow \Zc \rightarrow \Ze$, and $Y \rightarrow \Zc$, $e \rightarrow \Zc$ is not allowed.}
    \label{fig:graph_no_ey_x}
\end{figure}
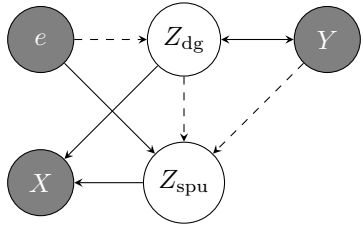

\subsection{On Valid DAGS:}
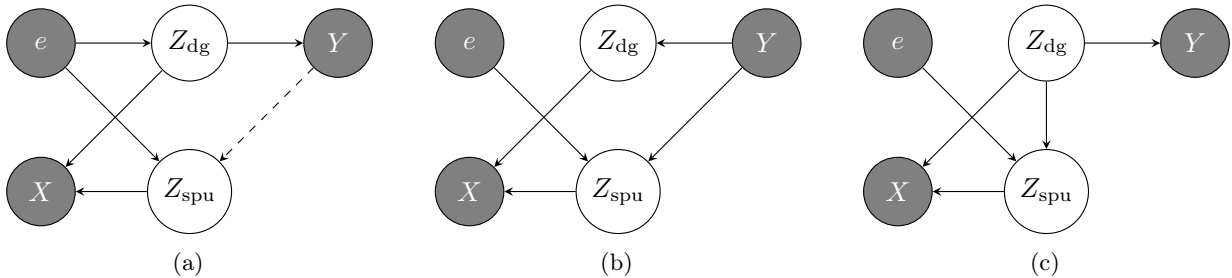
\begin{figure}[h]
    \centering
    \begin{subfigure}{0.31\textwidth}
        \resizebox{\textwidth}{!}{
            \input{files/graph_1.tex}
        }
    \caption{} \label{fig:app_first}
    \end{subfigure}
    \hfill
    \begin{subfigure}{0.31\textwidth}
        \resizebox{\textwidth}{!}{
            \input{files/graph_2.tex}
        }
    \caption{} \label{fig:app_second}
    \end{subfigure}
    \hfill
    \begin{subfigure}{0.31\textwidth}
        \resizebox{\textwidth}{!}{
            \input{files/graph_3.tex}
        }
    \caption{} \label{fig:app_third}
    \end{subfigure}
    \caption{{\bf Generative Processes.} Graphical model depicting the structure of our data-generating process - shaded nodes indicate observed variables. $X$ represents the observed features, $Y$ represents observed targets, and $e$ represents domain influences. There is an explicit separation of domain-general $\Zc$ and domain-specific $\Ze$ features combined to generate observed $X$. Dashed edges indicate  the possibility of an edge.}
    \label{fig:app_graph_other}
\end{figure}

We consider other edges that could be introduced to Figure \ref{fig:graph_no_ey_x} where $\Zc \not \indep \Ze \,|\, Y, e$, $\Ze \not \indep Y \,|\, \Zc$, or are not included in Figure \ref{fig:app_graph_other}. We then show that these edges either make the problem intractable or require new assumptions about the generative process -- note we do not discuss edges that induce a cycle, thus, are invalid.
\begin{enumerate}[label=(\roman*)]
    \item $e-Y$: we cannot have a direct edge in either direction $e$ between $Y$ otherwise, $Y$ is always dependent on $e$ and the problem becomes intractable.
    \item $e-X$: we cannot have a direct edge from $e-X$ without making additional parametric assumptions about the role of $e$ in $\Gamma(\Zc, \Ze, e)$.
    \item $\Ze \rightarrow Y$: we cannot have both $\Zc \rightarrow Y$ and $\Ze \rightarrow Y$, since then, both mechanisms are domain general. WLOG, we let $\Ze$ denote the features that never have domain-general mechanisms to $Y$.
    \item $Y \rightarrow \Zc \rightarrow \Ze$ and $Y \rightarrow \Zc \leftarrow e$: conditioning on $\Zc$ and/or $\Ze$ make $Y$ dependent on $e$, so $Y$ is always dependent on $e$ and the problem becomes intractable.
\end{enumerate}

\begin{table}[h!]
    \centering
    \caption{Generative Processes and Sufficient Conditions for Domain-Generality}
    \renewcommand{\arraystretch}{1.1}
    \label{apptab:easy_hard_DAGS}
    \resizebox{0.5\textwidth}{!}{
    \begin{tabular}{c|c|c|c}
        & \multicolumn{3}{c}{Graphs in Figure \ref{fig:app_graph_other}} \\
        \hline
        & (a) & (b) & (c) \\
        \hline
         $\Zc \indep \Ze \,|\, \{Y, e\}$ & \cmark & \cmark & \xmark \\
         Identifying $\Zc$ is necessary & \cmark & \cmark & \xmark \\
    \end{tabular}
}
\end{table}

\begin{table}[h!]
    \centering
    \caption{Generative Processes and Sufficient Algorithms}
    \label{tab:my_label}
    \begin{tabular}{c|c|c|c}
        & \multicolumn{3}{c}{Graphs in Figure \ref{fig:app_graph_other}} \\
        \hline
        {} & (a) & (b) & (c) \\
        \hline
         Solved by ERM & \xmark & \xmark & \cmark \\
         Solved by TCRI & \cmark & \cmark & \cmark
    \end{tabular}
\end{table}

\subsection{Fully Informative Invariant Features} \label{sec:app_fiif}
We briefly summarize \cite{ahuja2021invariance}'s results on minimax-optimality of Empirical Risk Minimization in the Fully Informative Invariant Features setting (their Lemma 4). First, we informally state their assumptions.

\begin{itemize}[label={}]
    \item Assumption 2: Linear structural equation model.
    \item Assumption 3-4: Bounded Features.
    \item Assumption 8: $\wc$ partitions $\Zcal$ up to noise $\eta_Y$.
\end{itemize}

These assumptions are implied by our Assumption \ref{assum:generative}.

\subsubsection{Proof Sufficiency of ERM \citep{ahuja2021invariance}}
If Assumptions 2, 4, and 8 hold, then there exists a classifier that puts a non-zero weight on the spurious feature and continues to be Bayes optimal in all the training environments.

\begin{proof}
    Choose an arbitrary non-zero vector and derive a bound on the margin of ($\wc,\, \gamma$), where $\wc$ is the true (optimal) linear predictor of $Y$ from $\Zc$. Recall domain-general and domain-specific features $\zc \in \Zcalc,\, \ze \in \Zcale$, respectively. Let $y^* = \sgn(\wc\cdot \zc)$. The margin of ($\wc,\, \gamma)$) at point $(\zc,\, \ze)$ with respect to $y^*$ is defined as:
    $$y^*(\wc\cdot \zc) + y^*(\gamma\cdot \ze).$$

    Using Cauchy-Schwartz inequality, we get $$|y^*(\gamma\cdot\ze)|  = |\gamma\cdot\ze| \le \|\|\gamma\|\ze\|.$$

    Since $\Ze$ is bounded, one can set $\gamma$ sufficiently small enough to control $y^*(\gamma \cdot \Ze)$. If $\|\gamma\| \le \frac{c}{2z^{\sup}}$, then $|\gamma\cdot\ze| \le \frac{c}{2}$, where $z^{\sup}$ satisfies that $\|z\| \le z^{\sup} \forall z \in \Zcale$. From Assumption 8, $\exists\, c > 0$ s.t., $$y^*(\wc\cdot\zc) \ge c.$$

    Using $|\gamma \cdot \ze| \le \frac{c}{2}$, the margin becomes $$y^*(\wc\cdot \zc) + y^*(\gamma\cdot \ze) \ge c - |\gamma\cdot \ze| \ge \frac{c}{2}.$$

    From the above equation, it follows that $\sgn\big((\wc, \gamma)\cdot(\zc, \ze)\big) = \sgn\big((\wc, 0)\cdot(\zc, \ze)\big) \forall \zc \in \Zcalc,\, \ze \in \Zcale$.

    Now, this condition is used to compute the error of a spurious classifier, i.e., based on $(\sc, \gamma)$. Define $\gee = I  \circ (\wc, \gamma) \circ \Gamma^{-1}$, where $I(\cdot)$ is an indicator function that returns 1 if its input is $\ge$ 0. The error achieved by $\gee$ is

    \begin{align*}
        R^e(\gee) &= \EE\big[Y^e \oplus I((\wc, \gamma)\cdot(\zc, \ze)\big] \\
        &= \EE\Big[I\big((\wc, 0)\cdot(\zc, \ze)\big)\oplus \eta_y \oplus I\big((\wc, \gamma)\cdot(\zc, \ze)\big)\Big] \\
        &= \EE[\eta_y].
    \end{align*}

    The error achieved by $\gee$ is then due to the noise in observed $Y$ and is, therefore, optimal in all domains.
\end{proof}

It follows from above that since $\gee$ is Bayes optimal in every domain, it is also the empirical risk minimizer (ERM) as it minimizes the sum of risks across training domains.

\FloatBarrier
\section{Proof of Proposition \ref{prop:main}}\label{sec:app_theory}

Assume that $\Phic(X)$ and $\Phie(X)$ are correlated with $Y$. Given Assumptions \ref{assum:generative}-\ref{assum:corr} and a representation $\Phi = \Phic \oplus \Phie$ that satisfies TIC, $\Phic(X) = \Zc \iff$ $\Phi$ satisfies TCRI.

\begin{proof}
    `only if'. Assume that $\Phic(X) = \Zc$. By the Total Information Criterion, we have that $\Phie(X) = \Ze$. We observe the following paths from $\Zc$ to $\Ze$: (i) $\Zc \rightarrow Y \rightarrow \Ze$, (ii) $\Zc \leftarrow e \rightarrow \Ze$, and  (iii) $\Zc \rightarrow X \rightarrow \Ze$. Conditioning on $Y, e$ blocks both paths (i) and path (ii); path (iii) contains a collider ($\Zc \text{ and } \Ze$ are common causes of $X$), so this path is blocked when $X$ is not in the conditioning set. So, $\Ze \indep \Zc \,|\, Y, e$ and therefore $\Phic(X) \indep \Phie(X) \,|\, Y, e$, which completes this direction.

    `if'. Assume that $\Phi$ satisfies TCRI. We proceed by contradiction. Let $\Phi = [\Phic; \Phie]$. We consider the following scenario for $\Phic \ne \Zc$.

    {\em Scenario 1 (causal aggregation)}: Assume that $\Phic(X) \subset \Zc$. From TIC, we have that $\Zc^\dag \subset \Phie(X)$, where $\Zc^\dag \subset \Zc$ is the subset of $\Zc$ not captured by $\Phic$. Since $\Phic(X)$ and $\Zc^\dag$ are colliders on $Y$, given both are subsets of $\Zc$, $\Phic(X) \not\indep \Phie(X) | Y, e$, violating TCRI and giving a contradiction. So, $\Zc \subset \Phi(X)$

    {\em Scenario 2 (anticausal exclusion)}: Assume that $\Phic(X) \subset \Ze$. From TIC, we have that $\Ze^\dag \subset \Phie(X)$, where $\Ze^\dag \subset \Ze$ is the subset of $\Ze$ not captured by $\Phic$. From Assumption \ref{assum:corr} (faithfulness), we have that $\Phic(X) \not\indep \Phie(X) | Y, e$, violating TCRI and giving a contradiction. So, $\Ze \not\subset \Phic(X)$.

    Combining scenarios 1-2, it follows that $\Phic(X) = \Zc$.
\end{proof}